\RequirePackage{fix-cm}
\documentclass[smallextended]{svjour3}       
\smartqed  
\usepackage{graphicx}
\usepackage{mathptmx}      
\usepackage{ifthen}
\usepackage{amssymb}
\usepackage{amsmath}
\usepackage{subfig}
\usepackage{adjustbox}
\usepackage{tikz}
\usepackage{hyperref}
\usepackage{color}

\usepackage{natbib}
\usepackage{url}
\usepackage[ruled,lined,titlenumbered,linesnumbered]{algorithm2e}
\newcommand{\entrances}{\ensuremath{\varepsilon}}
\newcommand{\objectives}{\ensuremath{\phi}}
\newcommand{\detectors}{\ensuremath{\delta}}
\newcommand{\blocked}{\ensuremath{\varpi}}
\newcommand{\cellSz}{\ensuremath{\zeta}}
\newcommand{\radius}{\ensuremath{\tau}}
\newcommand{\neutral}{\ensuremath{\theta}}

\newcommand{\Greedy}{Greedy}
\newcommand{\HC}{Hill Climbing}
\newcommand{\EA}{Evolutionary Algorithm}
\newcommand{\umda}{Univariate Marginal Distribution Algorithm}

\newcommand{\GRASP}[1]{GRASP$_{\alpha={#1}}$}
\newcommand{\GRASPHC}[1]{GRASP$_{\alpha={#1}}$+HC}
\newcommand{\UMDA}[1]{UMDA$_{\alpha={#1}}$}

\newcommand{\refRange}[3]{{#1}~\ref{#2}--\ref{#3}}
\newcommand{\refAndPage}[2]{{#2}~\ref{#1}}
\newcommand{\refTable}[1]{\refAndPage{#1}{Table}}
\newcommand{\refFig}[1]{\refAndPage{#1}{Figure}}
\newcommand{\refAlg}[1]{\refAndPage{#1}{Algorithm}}
\newcommand{\refFigs}[2]{\refRange{Figures}{#1}{#2}}

\newcommand{\refEq}[1]{\refAndPage{#1}{Equation}}

\newcommand{\refSect}[1]{\refAndPage{#1}{Section}}

\DeclareMathOperator*{\argmin}{arg\,min}
\DeclareMathOperator*{\argmax}{arg\,max}

\newcommand{\kw}[1]{\textbf{#1}}
\newcommand{\fnctn}[1]{\textsc{#1}}

\journalname{Journal of Heuristics}
\begin{document}

\title{Metaheuristic Approaches to the Placement of Suicide Bomber Detectors
\thanks{Authors acknowledge support from Spanish Ministry of Economy and Competitiveness
and European Regional Development Fund (FEDER) under project EphemeCH
(TIN2014-56494-C4-1-P).}

\thanks{This document is a preprint of Cotta, C., Gallardo, J.E. Metaheuristic approaches to the placement of suicide bomber detectors. J Heuristics 24, 483–513 (2018). \url{https://doi.org/10.1007/s10732-017-9335-z}}

}


\author{Carlos Cotta         \and
        Jos\'{e}~E.~Gallardo 
}


\institute{Carlos Cotta         \and
	Jos\'{e}~E.~Gallardo \at
              Department {\em Lenguajes y Ciencias de la Computaci\'on},
              Universidad de M\'alaga,\\
              Campus de Teatinos, 29071 - M{\'a}laga, Spain. \\
              Tel.: +34 95 213 7158\\
              Fax: +34 95 213 1397\\
              \email{\{ccottap,pepeg\}@lcc.uma.es}           
}

\date{Received: date / Accepted: date}

\maketitle

\begin{abstract}
Suicide bombing is an infamous form of terrorism that is becoming increasingly prevalent in
the current era of global terror warfare. We consider the case of targeted attacks of this kind,
and the use of detectors distributed over the area under threat as a protective countermeasure.
Such detectors are non-fully reliable, and must be strategically placed in order to maximize
the chances of detecting the attack, hence minimizing the expected number of casualties. 
To this end, different metaheuristic approaches based on local search and on population-based 
search are considered and benchmarked against a powerful greedy heuristic from the literature.
We conduct an extensive empirical evaluation on synthetic instances featuring very diverse
properties. Most metaheuristics outperform the greedy algorithm, and a hill-climber
is shown to be superior to remaining approaches. This hill-climber is subsequently subject to
a sensitivity analysis to determine which problem features make it stand above the greedy
approach, and is finally deployed on a number of problem instances built after realistic
scenarios, corroborating the good performance of the heuristic.


\end{abstract}

\section{Introduction} \label{intro}

\begin{figure}[t!] \begin{center}
\includegraphics[angle=0,width=\textwidth]{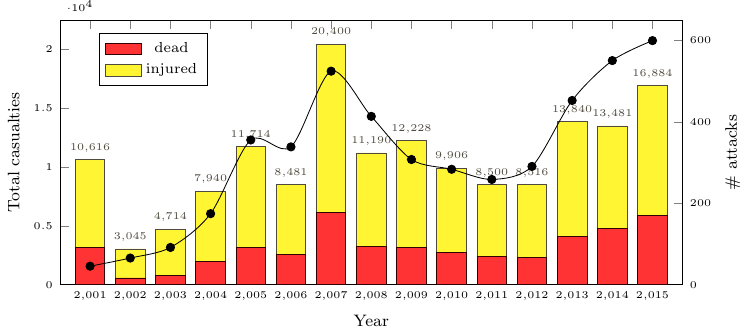}
\caption{The bart chart shows the number of casualties (dead and
injured) in suicide attacks in the period 2001-2015 (total year
figures on top of each bar, scale on the left). The black curve
indicates the number of suicide attacks in this time frame
(scale on the right). Source: own elaboration based on data from
\citep{cpost16database}.} \label{fig:attacks} \end{center}
\end{figure}

At the time of writing this paper, the Atat\"urk Airport in
Istanbul was subject to a terrorist attack whose perpetrators
carried automatic weapons and explosive belts. As a result of
the shooting and the detonation of the suicide bombs, 44 people
were killed (in addition to the three terrorists) and more than
200 were injured \citep{kukuriz16istanbul}. A few days afterwards,
a suicide car bombing killed 292 people and left more than
200 people wounded in a shopping district of Baghdad,
in what constituted the deadliest single bomb attack in Iraq since
2007 \citep{afp16death}.
These are only two recent examples of suicide bombings, a form
of terrorism that not only causes about four time more
casualties than other kinds of terrorist attack
\citep{rand09database}, cf. \citep{hoffman03logic} but also
instills a sense of fear in the society as a whole that
undermines public confidence in the authorities and contributes
to subjugate those living under this threat
\citep{hoffman03logic}. As a result of the relative
inexpensiveness (they do not require escaping logistics nor
sophisticated equipment for remote operation) and effectiveness
of this kind of attacks, they have become increasingly
prevalent; see \refFig{fig:attacks}. Indeed, they are not
just very frequent in conflicting areas such as the Middle East
--1,192 attacks in the period 1982-2015, only counting those
conducted using explosive belts \citep{cpost16database}-- but
also constitute a global threat that has caused nearly 50,000
deaths worldwide in the 21st Century, with an average of about
10 fatalities and 24 non-fatal casualties per attack (counting
all suicide attacks -- data from \citet{cpost16database}).

In response to this ongoing threat, security forces and
intelligence agencies are strengthening their efforts.
Obviously, the nature of this kind of attack makes it infeasible
to rely on standard deterrence measures based on direct
retaliation. Hence, other members of the terrorist network
providing intellectual or logistic support to the attacker (and
in general any assets valued by the latter) are targeted instead
\citep{kroenig12deter}. In any case, this does not fully deters
this kind of attacks, so they must be also fought against by
trying to deny their benefits, either at a strategic level
(ensuring that the ultimate goals of the terrorists will not be
achieved even if the attack was successful) or at a tactical
level (trying to make the attacks unsuccessful). Heightened
security measures in airports, government buildings or military
installations are examples of this latter kind of
counterterrorism measures.

A potential problem of these explicit, partly invasive security
measures is that they cannot be readily exported to other kind
of environments also susceptible to terror attacks (e.g.,
placing airport-like scanners surrounding a city square or busy
shopping street, or frisking every passer-by is out of
question). More subtle surveillance can be accomplished by using
chemical or biological sensors tailored to detect traces of
explosives in their proximity \citep{NRC04existing,singh07sensors,Yinon2007, caygill12current}.
In this sense, \citet{kaplan05operational} studied the deployment of such
sensors in urban areas subject to pedestrian suicide-bombing
attacks. They analyzed different scenarios (an urban grid of
streets and a large plaza or park) and concluded that, while
such sensors are not likely effective against random attacks,
they can play an important role in the defense of known targets.
Building on this work, \citet{Xiaofeng2007} considered the case
of a threat area with known targets on which a number of
non-fully reliable sensors are deployed. They proposed a
branch-and-bound algorithm (BnB) and a greedy constructive
heuristic to determine the location of these sensors so as to
minimize the casualties. The BnB algorithm does not scale well
with problem size, but the greedy heuristic is relatively
effective (although it will not guarantee finding the optimal
solution in general).

Following these previous results, the problem is here approached
for the first time to the best of our knowledge
from the point of view of iterative heuristics and metaheuristic
techniques. We propose several algorithms based on local search,
greedy randomized adaptive search procedures and evolutionary
algorithms, and analyze their performance, comparing it to the
greedy approach previously mentioned on an extensive collection
of synthetic and real scenarios. Before presenting the
algorithms considered and the experimental setup, let us firstly
define formally the problem tackled. This is done in next
section.

 \section{Problem Description}

\begin{figure}[t!] 
\begin{center}
		\adjustbox{trim={0\width} 0 {0\width} 0,clip}
		{
		\includegraphics[angle=0,scale=0.8]{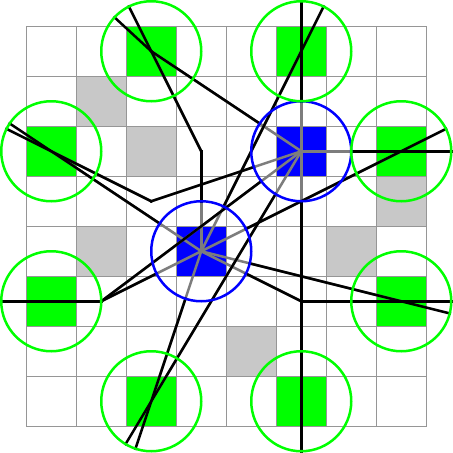}
		}
\caption{An $8 \times 8$ scenario. Green cells correspond to
entrances, blue ones to objectives and gray cells represent
blocked locations. Lines are shortest paths from each entrance
to each objective (note that paths would be extended backwards should
a detector be placed on any of the entrances). 
The dimension of each cell is 10m $\times$ 10m,
and the attacker must be neutralized at least 10m away
from an objective (we assume that the terrorist moves at a speed
of 1m/s and that aborting the attack requires no more than 10
seconds). } \label{fig:map3} 
\end{center} 
\end{figure}

Let us assume the goal is to protect a certain threat area or
scenario which we can model as a rectangular grid
${A}=\{A_{ij}\}_{m\times n}$. Each cell in this grid represents a
small square subarea which can be \emph{blocked} if there is some
physical obstruction (a wall, a monument, etc.) that precludes
its being traversed by the attacker, or \emph{unblocked} if it
can be freely accessed. Some of these latter unblocked cells
can be also \emph{objectives}, that is, cells that contain
threatened individuals who may be targeted by the attacker. To
carry on this threat, the attacker can enter the scenario
through some \emph{entrances}, namely specific unblocked cells
typically placed in the boundaries of the grid. From any of
those entrances, the attacker will try to reach any of the
objectives following always a shortest path. Following
\citep{Xiaofeng2007}, the attacker is able to move in a straight
line from the center of a grid cell to the center of any other
one for which this line does not intersect with a blocked cell.
Thus, the path followed by the attacker will be composed by a
number of straight segments, and can be computed by running
classical algorithms --such as, e.g., Dijkstra's algorithm or
A*-- on a weighted complete graph $G(V,E,w)$, where
\begin{itemize} 
\item $V=\{\langle i,j\rangle\ |\ A_{ij}$ is unblocked$\}$, i.e., a 
vertex per unblocked cell, 
\item $E = V\times V$, and \item $w:E\rightarrow\mathbb{R}$ 
is defined as
\begin{equation} 
w((\langle i_1,j_1\rangle, \langle i_2,j_2\rangle))= \cellSz\sqrt{(i_1-i_2)^2+(j_1-j_2)^2},
\end{equation} 
(where \cellSz\ is the side of each grid cell) if
there is no obstacle in the straight path among 
both cells, and
$w(E)=\infty$ otherwise. 
\end{itemize} 
\refFig{fig:map3} shows an example $8\times8$ scenario 
with 8 entrances, 6 blocked
cells and 2 objectives. Notice how paths try to follow always a
straight line towards the objective, taking small detours if
there are obstacles on the way.

With this setting in mind, the problem under consideration
amounts to placing some \emph{detectors} in 
different unblocked cells of the grid. These
detectors are perfectly concealed from the attacker, but they
are not fully reliable and will only detect attackers that
travel within a certain detection radius \radius\ of the
detector with a probability $p$ that depends on the length of
the attacker's path inside this detection area (a circle of
radius \radius\ centered in the detector's location). Detection
does not imply neutralization of the attacker though. It only
amounts to firing some alarm that elicits response from some
enforcing agents, and that can lead to effective negation of the
attack only if the attacker is not yet very close to the
objective, and with some probability \neutral. Let us formalize
the whole process in the following.

Let \entrances\ and \objectives\ be the number of entrances and
objectives respectively. For each entrance $e_i$, $1\leqslant i
\leqslant \entrances$, and objective $o_j$, $1\leqslant
j\leqslant\objectives$ there is a path $P_{ij}$ (we assume for
simplicity that the shortest path is unique -- considering
multiple paths does not fundamentally affect the analysis
below). The attacker will pick an entrance $e_i$ and an
objective $o_j$ with probability $\gamma_{ij}$ (obviously,
$\sum_{i=1}^\entrances\sum_{j=1}^\objectives\gamma_{ij} = 1$)
and will subsequently move along $P_{ij}$ at speed $v$.
Neutralizing the attack once detected takes some time $t_n$ --at
least 10s according to \citet{kaplan05operational}-- and hence
once the attacker is at distance $vt_n$ from the objective
(distance measured along the path $P_{ij}$) no effective
neutralization is possible. Therefore, let us define
$\bar{P}_{ij}$ as the portion of the path $P_{ij}$ outside this
``dead'' zone, in which timely detection is still possible.

\begin{figure}[t!] 
\begin{center}
		\includegraphics[angle=0,width=0.5\textwidth]{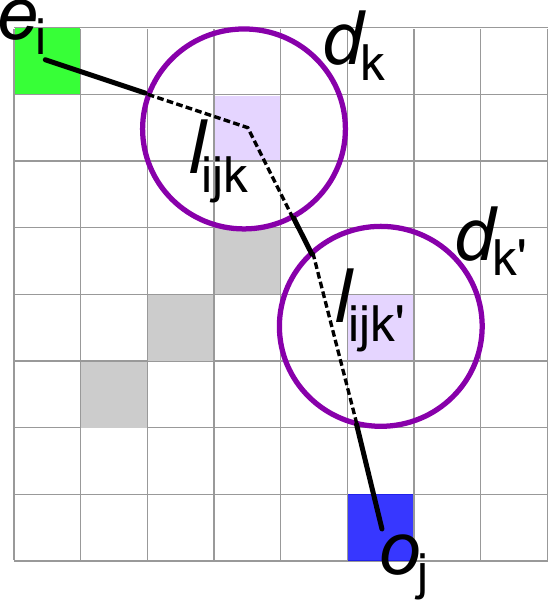}		
		\caption{A shortest path going from entrance $e_i$ to objective $o_j$. Areas monitored by detectors $d_k$ and $d_k'$ are those enclosed by circumferences. The segments of this path detected by both detectors are shown with a dotted line and denoted by $l_{ijk}$ and $l_{ijk'}$.} \label{fig:paths} 
\end{center}
\end{figure}

Now, let \detectors\ be the number of detectors in the scenario.
A certain detector $d_k$, $1\leqslant k\leqslant\detectors$,
will timely detect an attacker traveling through $P_{ij}$ with
probability 
\begin{equation} 
p_{ijk} = 1-\exp(-\eta l_{ijk})
\end{equation} 
where $l_{ijk}$ is the length of the segment of
$\bar{P}_{ij}$ within the circle of radius \radius\ centered at
$d_k$, and $\eta>0$ is a parameter (the detector's instantaneous
detection rate). It follows that the larger this segment, the
larger the detection probability (and obviously $p_{ijk}=0$ if
$l_{ijk}=0$, i.e., if the path is outside the detection radius
of $d_k$) -- see \refFig{fig:paths} for an example. 
Following \citep{Xiaofeng2007}, detectors are assumed
to work independently of one another, and therefore the total
probability of non-detection $\tilde{D}_{ij}$ if the attacker
follows $P_{ij}$ is 
\begin{equation} 
\tilde{D}_{ij} =\prod_{k=1}^\detectors (1-p_{ijk}) = \prod_{k=1}^\detectors \exp(-\eta l_{ijk}) = \exp(-\eta \sum_{k=1}^\detectors l_{ijk})
\end{equation} 
In case of non-detection, the attacker will reach
the objective $o_j$ causing a number of casualties $C_j$. This
will be also the case, should the attacker be timely
detected but not effectively neutralized (an event that will
happen with probability $1-\neutral$). The expected number of
casualties $W_{ij}$ for this particular path will thus be 
\begin{equation} 
W_{ij} = \tilde{D}_{ij}C_j + (1-\tilde{D}_{ij})(1-\neutral)C_j = C_j\left[\tilde{D}_{ij}\neutral + (1-\neutral)\right] 
\end{equation}
The total number of casualties will take into account all
possible paths $P_{ij}$ the attacker can take, that is, between
any entrance $e_i$ and objective $o_j$ (and recall each of these
is picked with probability $\gamma_{ij}$): 
\begin{eqnarray} 
W & = &\sum_{i=1}^\entrances\sum_{j=1}^\objectives \gamma_{ij}W_{ij} = \sum_{i=1}^\entrances\sum_{j=1}^\objectives \gamma_{ij}C_j\left[\tilde{D}_{ij}\neutral + (1-\neutral)\right] = \\ 
   & = & \sum_{i=1}^\entrances\sum_{j=1}^\objectives \gamma_{ij}C_j(1-\neutral) +
\sum_{i=1}^\entrances\sum_{j=1}^\objectives \gamma_{ij}C_j\tilde{D}_{ij}\neutral= \\ 
  & = & (1-\neutral)\sum_{i=1}^\entrances\sum_{j=1}^\objectives \gamma_{ij}C_j + \neutral\sum_{i=1}^\entrances\sum_{j=1}^\objectives \gamma_{ij}C_j\exp(-\eta \sum_{k=1}^\detectors l_{ijk})\label{eq:fitness} 
\end{eqnarray}

The objective of the problem is thus placing \detectors\
detectors on the grid such that Equation (\ref{eq:fitness}) is
minimized (this actually implies minimizing the second term
therein since the first one is independent of the placement of
the detectors). We shall denote this problem as the
Optimal Placement of Suicide-Bomber Detectors (OPSBD).

\section{Description of Algorithms}
\label{sect:Algorithms}

In this section, we describe the different algorithms for the OPSBD problem 
that we have considered in this work. To this end, we shall start by presenting
some general algorithmic considerations that are applicable to all algorithms
considered, namely a cache data structure aimed to avoid recomputation and
a dominance criterion that reduces the search space. 
Subsequently, we will provide a detailed description of all heuristics, 
namely a Greedy Randomized Adaptive Search Procedure 
(GRASP), a Hill Climber (HC), an Evolutionary Algorithm (EA) and a \umda{} (UMDA).
Prior to these and for the sake of completeness, we will also describe 
the \Greedy{} algorithm proposed by \citet{Xiaofeng2007}, which will be
used to benchmark the remaining procedures.

\subsection{General Algorithmic Considerations}
\label{sect:cache}
All algorithms compared in this work use a cache data structure in order to 
accelerate the otherwise repetitive computations needed during the optimization 
of a particular instance.

This memory consists of several components. The first one is a three-dimensional 
array ${\Lambda}=\{\Lambda_{ijp}\}_{m\times n\times r}$ whose first two dimensions correspond 
to the number of rows ($m$) and columns ($n$) of the map. The third one corresponds 
to the total number of paths going from each entrance in the map to each objective ($r=\entrances\objectives$). 
Each entry in this data structure stores the length of the segment of the path that 
would be detected by placing a detector at the center of the cell denoted by the 
two first dimensions, i.e., if detector $d_k$ is placed at cell $(i',j')$, $\Lambda_{i'j'p} = l_{ijk}$,
where $P_{ij}$ is the $p$th path under a certain predefined order and $l_{ijk}$ is as defined
in previous section. 

\begin{figure}[t!] \begin{center}
\includegraphics[angle=0,scale=0.8]{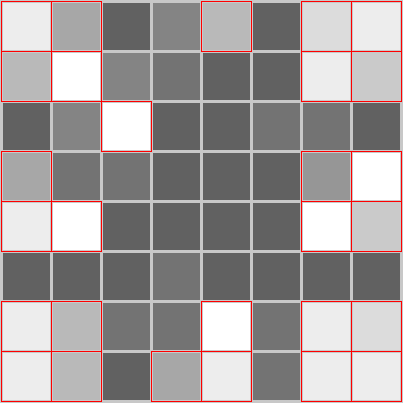}
\caption{Dominated cells in the map from \refFig{fig:map3}, assuming that a total of  $\detectors=3$ detectors are to be placed. The detector effectiveness radius is $\radius = 10$m. Red bordered
cells can be safely disregarded as candidate locations for placing
the detectors. Lightest cells are more dominated than darker ones.}
\label{fig:map1:dominances} \end{center} \end{figure}

Another component of the cache is a  bi-dimensional array ${\Delta}=\{\Delta_{ij}\}_{m\times n}$,
which stores the number of 
times each cell in the map is dominated by remaining cells. We say that a cell $c_1 = (i_1,j_1)$ 
is dominated by another cell $c_2=(i_2,j_2)$ if, 
\begin{eqnarray}
\forall p\in{1,\dots \entrances\objectives}:\quad  & \Lambda_{i_1j_1p} \leqslant \Lambda_{i_2j_2p} \\
\exists p\in{1,\dots \entrances\objectives}:\quad & \Lambda_{i_1j_1p} < \Lambda_{i_2j_2p}
\end{eqnarray}
that is, for all paths in the map, the length of the segment of that path that would be 
detected by placing a detector at cell $c_1$ is less than or equal to the one detected by 
placing instead the detector at cell $c_2$, and, for at least one path, the path detected 
by a detector at $c_1$ is strictly less than that detected from $c_2$. This concept of 
dominance can be used to prune the search space of the problem since those cells that 
are dominated by a number of cells greater than or equal to the total number of detectors 
to be placed at the map (i.e., $\Delta_{ij}\geqslant\detectors$) can be pruned, since there 
exists at least \detectors\  cells better than those. \refFig{fig:map1:dominances} 
illustrates dominance values for the scenario shown before.

\subsection{Greedy Algorithm}
\label{sect:GREEDY}
As mentioned above, this algorithm (denoted as \Greedy{} henceforth) corresponds to the one originally described 
in \citep{Xiaofeng2007} and will be used to establish a baseline for comparison 
purposes. This greedy heuristic is a constructive method that places the \detectors\ 
detectors one at a time as depicted in the pseudo-code provided in \refAlg{fig:alg:GREEDY}. 

Firstly, the set of candidate detectors is initialized with all non-blocked cells in the 
map minus those that are dominated by at least \detectors\ cells (line 1). The algorithm 
starts from an empty solution (line 2) and from there on, the \detectors\ detectors 
are successively added to the current partial solution $\mathit{sol}$ (lines 3-15). 
In each iteration, the candidate detector whose 
incorporation leads to the best extended solution --i.e., locally minimizing Equation 
(\ref{eq:fitness})-- is selected (lines 6-12), added to current solution (line 13) and removed from the set 
of candidate detectors (line 14).

As it can be seen, this greedy algorithm performs different locally optimal decisions 
which can be taken with a low computational cost. However, due to the myopic functioning 
of this scheme, the resulting final solution is not guaranteed to be optimal. Although 
it was shown in \citep{Xiaofeng2007} that the quality of solutions provided by the greedy 
algorithm for this particular problem is not very far from the global optimum for the 
relatively small-sized instances that were considered in that work, it will be later 
shown in \refSect{sec:results} that other heuristics can provide better 
solutions for this problem if larger instances are considered.

	\begin{algorithm}[!t]
		\caption{Greedy Algorithm\label{fig:alg:GREEDY}}
		\KwIn{\detectors\ (number of detectors to be placed)}	
		$\mathit{candidates} := \{(r,c)\ |\ 1 \leqslant r \leqslant m, 1 
		\leqslant c \leqslant n, A_{rc}$ is unblocked $, 
		\Delta_{r,c}<\detectors \} $\;	
		$\mathit{sol} := \varnothing$\;
		\While{$\mathit{length}(\mathit{sol}) < \detectors$}
		{	
			$\mathit{fitness}^* := \infty$\;
			$\mathit{detector}^* := \mathrm{null}$\;
			\For {$d \in \mathit{candidates}$}
			{
				$\mathit{tentative} := \mathit{sol} + \{ d \}$\;
				\If{$\mathit{value}(\mathit{tentative}) < \mathit{fitness}^*$}
				{
					$\mathit{detector}^* := d$\;
					$\mathit{fitness}^* := \mathit{value}(\mathit{tentative})$\;
				}		 
			}
			$\mathit{sol} := \mathit{sol} + \{\mathit{detector}^* \}$\;
			$\mathit{candidates} := \mathit{candidates} \setminus 
			\{\mathit{detector}^*\}$\;
		}	
		\Return{$\mathit{sol}$}\;
	\end{algorithm}

\subsection{Greedy Randomized Adaptive Search Procedure}
\label{sect:GRASP}
The Greedy Randomized Adaptive Search Procedure (GRASP) \citep{GRASP} is a metaheuristic that tries to alleviate the myopic functioning associated with greedy algorithms. This is also a constructive algorithm that adds one component of the solution at the time, but unlike the greedy algorithm, the one leading to largest local improvement is not always selected. 

The pseudo-code in \refAlg{fig:alg:GRASP} is a description of the specific incarnation 
of the general GRASP scheme that we have considered in this work. First of all, a {\em candidate list} ($\mathit{CL}$) of partial solutions is built by extending the current  solution ($\mathit{sol}$) 
with all candidate detectors (lines 4-8). Subsequently, a {\em restricted candidate list} 
($\mathit{RCL}$) is built by selecting some detectors from the candidate list 
(lines 9-17). This is done by using 
a threshold for the quality of tolerable partial solutions ($\mu$), which we compute  
as a percentage of the difference between the values of the best and 
worse solutions in the candidate list (lines 9-11). Parameter $\alpha$ ($0 \leqslant \alpha \leqslant 1$) 
of the algorithm controls the degree of greediness used. For instance, if $\alpha$ is 0, 
then only the best extended solution is included in the $\mathit{RCL}$ (this 
would correspond to a 
greedy selection strategy). For greater values of $\alpha$, more tentative decisions can 
be included in the $\mathit{RCL}$ (all of them would be included when $\alpha = 
1$) so that selection 
is less intense and the search is diversified. The next detector to be added to the solution 
is randomly chosen from the $\mathit{RCL}$ in an uniform way (line 18). This 
process concludes when 
the solution being built is complete, i.e., it includes \detectors\ detectors. Although it is not 
shown in the pseudo-code, this whole process can be repeated until the maximum execution
time is exhausted. The best solution found is then returned.

	\begin{algorithm}[!t]
		\caption{GRASP Algorithm	\label{fig:alg:GRASP}}
		\SetKwInOut{Input}{Input}  
		\Input{\detectors \ (number of detectors)\\
				$\alpha$ (greedy temperation parameter)}
		 $\mathit{candidates} := \{(r,c)\ |\ 1 \leqslant r \leqslant m, 1 \leqslant c \leqslant n, A_{rc}$ is unblocked$, \Delta_{rc}<\detectors \} $\;
		 $\mathit{sol} := \varnothing$\;
		 \While{$\mathit{length}(\mathit{sol}) < \detectors$}
		 {		
			  $\mathit{CL} := \varnothing$\;
			  \For{$d \in \mathit{candidates}$}
			  {
			   	$\mathit{tentative} := \mathit{sol} + \{ d \}$\;
			   	$\mathit{CL} := \mathit{CL} \cup \{(\mathit{tentative}, d) \}$\;
			  }
			  $\mathit{min} :=\argmin_{(s,d) \in \mathit{CL}}{\mathit{value}(s)}$\;
			  $\mathit{max} :=\argmax_{(s,d) \in \mathit{CL}}{\mathit{value}(s)}$\;
			  $\mu := \mathit{min} + \alpha (\mathit{max} - \mathit{min})$\;
			  $\mathit{RCL} := \varnothing$\;
			  \For{$(s,d) \in \mathit{CL}$}
			  {  
			  	\If{$\mathit{value}(s) \leqslant \mu$}
			  	{
			   		$\mathit{RCL} := \mathit{RCL} \cup \{(s,d) \}$\;
			   	}
			  }
			  $(\mathit{sol},d) := \fnctn{RandomSelect}(\mathit{RCL})$\;
	   	      $\mathit{candidates} := \mathit{candidates} \setminus \{ d \}$\; 
		 }	
		 \Return{$\mathit{sol}$}\;
	\end{algorithm}

\subsection{Hill Climbing}
\label{sect:HC}
A simple and effective class of optimization heuristics based on the
notion of {\em neighborhood} is Local Search
\citep{AartsLenstra97,HoosStutzleBook}.  As opposed to constructive 
methods, Local Search algorithms are procedures that work with complete solutions to the problem.
In these algorithms, an initial
solution is provided (which can either be constructed randomly or by means of another
heuristic). Then, its neighborhood is examined, and if a better
solution is found, a move is performed towards it. The process is
repeated with the new solution until a local optimum is found. 

Here, we consider a  Hill Climbing (HC) algorithm. In the pseudo-code for this procedure 
(shown in \refAlg{fig:alg:HC}), $sol$ denotes a solution to be improved, represented as a 
collection of locations for different detectors. The algorithm substitutes one detector location 
in the current solution for all unused candidate locations for the problem instance 
($\fnctn{Replace}(\mathit{sol},i,d)$ denotes the solution that is obtained by replacing $i$-th 
detector in solution $\mathit{sol}$ by alternative detector $d$ -- see line 7) and selects the one leading 
to a better solution (lines 8-11). If, as a result of this replacement, the new solution is better than the 
current one, the corresponding move is accepted (line 9) and an improvement is acknowledged (line 10). 
Using the new solution, the same process is repeated for the rest of detectors. The process 
is iterated as long as the solution has been improved, until no further enhancement is possible. 
The final solution is a local optimum and is returned as a result. Different 
starting solutions can 
be used and the whole scheme can be repeated until the maximum allowed execution time is reached.

	\begin{algorithm}[!t]
	\caption{Hill Climbing Algorithm	\label{fig:alg:HC}}
		\KwIn{$\mathit{sol}$ (a collection with the coordinates of \detectors\ detectors)}	
		$\mathit{candidates} := \{(r,c)\ |\ 1 \leqslant r \leqslant m, 1 \leqslant c \leqslant n, A_{rc}$ is unblocked$, \Delta_{rc}<\detectors \} $\;
		$\mathit{improvement} := \mathrm{true}$\;
		\While{$\mathrm{improvement}$}
		{
		  	$\mathit{improvement} := \mathrm{false}$\;
		  	\For{$i := 1$ \kw{to} $\mathit{length}(\mathit{sol})$}
		  	{
		   		\For{$d \in (\mathit{candidates} \setminus \{d\,|\,d \in \mathit{sol} \})$}
		   		{		
		    			$\mathit{tentative} := 
		    			\fnctn{Replace}(\mathit{sol},i,d)$\;
		    			\If{$\mathit{value}(\mathit{tentative}) < \mathit{value}(\mathit{sol})$}
		    			{		
		     				$\mathit{sol} := \mathit{tentative}$\;
		      				$\mathit{improvement} := \mathrm{true}$\;
		    			}
		   		}
		  	}
		 }	
		 \Return{$\mathit{sol}$}\;
	\end{algorithm}

\subsection{Evolutionary Algorithm}
\label{sect:EA}
Evolutionary algorithms (EAs) \citep{Eiben:2003:IEC} are black-box optimization 
procedures inspired by the biological evolution of species that work with a population 
of solutions subject to different operations such as reproduction, recombination, 
mutation and replacement.
In this section, we describe a steady-state evolutionary algorithm that we have 
considered in order to tackle the OPSBD problem.

The pseudo-code of the EA is depicted in \refAlg{fig:alg:EA}. 
As shown, the algorithm uses a population ($\mathit{pop}$) of $\mathit{popSize}$ 
non-repeated individuals, each one corresponding to one full solution to the problem. 
Each of these individuals is represented as a vector whose length is the number of 
detectors (\detectors) to be placed at the map. Hence, each element in the vector 
corresponds to one of the cells in the map where one of the detectors should be 
placed. Similarly to previous heuristics, the search space explored by the EA 
is restricted to  
non-blocked cells in the 
map minus those that are dominated by at least \detectors\ cells.
Each solution in the population is initialized by randomly selecting in an uniform way 
some of these cells (lines 1-4). Afterwards, the following process is repeated, 
until the 
maximum allowed execution time is reached:
\begin{itemize}
	\item With probability of selection $p_{X}$, two parents are selected from the 
	current population using {\em binary tournament selection} (lines 7-9). Let 
	us now consider the set comprising the union of detector placements 
	included in 
	selected 
	individuals. A new individual, constituting the offspring for this 
	generation, is then defined by sampling in a random way \detectors{} 
	different placements from this previous set.
	\item Otherwise (with probability $1-p_{X}$) a random individual of current 
	population is selected as the offspring (line 11).
	\item Each component of the offspring is mutated with probability $p_m$ by 
	replacing the corresponding detector in that component with another one not 
	included in the solution (line 13), and the resulting individual is evaluated (line 14).
	\item As to replacement, the worst individual in the current population is replaced by the offspring (line 15). 
\end{itemize}
Finally, when the execution time limit is reached, the best individual found  is returned as a solution.

	\begin{algorithm}[!t]
	\caption{Evolutionary Algorithm\label{fig:alg:EA}}
		\SetKwInOut{Input}{Input}  
		\Input{$\mathit{popSize}$ (population size)\\
			    $p_{X}$ (recombination probability)\\
			    $p_m$ (mutation probability)}	
		 \For{$i :=1$ \kw{to} $\mathit{popSize}$}
		 {
		 	 $\mathit{pop}_i :=$ \fnctn{RandomIndividual}$()$\;
		 	 \fnctn{Evaluate}($\mathit{pop}_i$)\;
		 }
		 \While{allowed runtime \kw{not} exceeded}
		 {
		 	 \eIf{recombination is performed($p_{X}$)}
		 	 {
		 	 	 $parent_1 :=$ \fnctn{Select}($\mathit{pop}$)\;
		 	 	 $parent_2 :=$ \fnctn{Select}($\mathit{pop}$)\;
		 	 	 $\mathit{offspring} :=$ \fnctn{Recombine}($parent_1$, $parent_2$)\;
		 	 }{
		 	 	 $\mathit{offspring}:=$ \fnctn{Select}($\mathit{pop}$)\;
		 	 }
		 	 $\mathit{offspring} :=$ \fnctn{Mutate}($p_m,\mathit{offspring}$)\;
 	 		\fnctn{Evaluate}($\mathit{offspring}$)\;
		 	 $\mathit{pop} :=$ \fnctn{Replace}($\mathit{pop}$, $\mathit{offspring}$)\;
		 }	
		 \Return{best solution found}\;
	\end{algorithm}

\subsection{\umda{}}
\label{sect:UMDA}
Finally, we have taken into account a \umda{} (UMDA) \citep{Muhlenbein1996}, a metaheuristic which belongs to the class of Estimation of Distribution Algorithms (EDAs) \citep{larranaga2002estimation}. EDAs are themselves based on Evolutionary Algorithms, but whereas most classical EAs build new solutions using recombination and mutation operators, EDAs do this by sampling an explicit probability model which is updated along the optimization process. This model can be encoded in different ways, and, in the case of UMDAs, a simple univariate linear model is used. The general functioning of an EDA is illustrated in \refAlg{fig:alg:UMDA}.

\begin{algorithm}[!t]
	\caption{Estimation of Distribution Algorithm\label{fig:alg:UMDA}}
	\KwIn{$\mathit{popSize}$ (population size)}	
	$\mathit{model} :=$ \fnctn{BuildInitialModel}$()$\;
	\For{$i :=1$ \kw{to} $\mathit{popSize}$}
	{
		$\mathit{pop}_i :=$ \fnctn{SampleIndividual}$(\mathit{model})$\;
		\fnctn{Evaluate}($\mathit{pop}_i$)\;
	}
	\While{allowed runtime \kw{not} exceeded}
	{
		$\mathit{pop'} :=$ \fnctn{Select}$(\mathit{pop})$\;
		$\mathit{model} :=$ \fnctn{BuildModel}$(\mathit{pop'})$\;
		\For{$i :=1$ \kw{to} $\mathit{popSize}$}
		{
			$\mathit{pop}_i :=$ \fnctn{SampleIndividual}$(\mathit{model})$\;
			\fnctn{Evaluate}($\mathit{pop}_i$)\;
		}
	}	
	\Return{best solution found}\;
\end{algorithm}

In our case, we use a GRASP-based encoding \citep{cotta04hybrid}, that is, individuals 
in the population encode the indexes in the RCL of the decisions that the 
GRASP algorithm would pick to place each detector, and 
the univariate 
probability distribution is used to model these indexes. More precisely, each 
solution is a vector
$\vec{x}\in\mathbb{N}^{\detectors-1}$, where $x_i$ indicates that the $i$th 
detector is picked as the $x_i$th best 
(according to the 
greedy selection criterion) option available at that point, once that the previous
$i-1$ detectors have been placed (e.g., vector ${\vec x}={\vec 0}$ would encode
the greedy solution described in \refSect{sect:GREEDY} since all detectors would
be picked according to the best local decision). Note that the 
length of $\vec{x}$
is $\detectors-1$ because it does not make sense to place the last detector in 
any other location
than the locally optimal one. 

In the univariate model these variables are
assumed to be independent and hence the probability distribution $p(\vec{x})$ is
factorized as

\begin{equation}
p(\vec x=\langle v_1,\dots,v_{\detectors-1}\rangle) = \prod_{i=1}^{\detectors-1} p(x_i=v_i).
\end{equation}
UMDA computes $p(x_i=v_i)$ as
\begin{equation}
p(x_i=v_i) = \frac{1}{|\mathit{pop'}|}\sum_{j=1}^{|\mathit{pop'}|} [pop'_{ji}=v_i],\label{eq:umda}
\end{equation}
where $pop'_{ji}$ is the $i$th variable of the $j$th solution in $pop'$ and $[\cdot]:\mathbb{B}\rightarrow\{0,1\}$ is an indicator function ($[\mathrm{true}]=1$, $[\mathrm{false}]=0$). 				

At the beginning of the algorithm, the model is initialized (\fnctn{BuildInitialModel} 
in line 1) in a heuristic way: the GRASP algorithm described in \refSect{sect:GRASP} is run $\mathit{popSize}$ times
for a certain fixed value of $\alpha$, and the decision vectors generated in those runs are used to create the initial model as indicated in \refEq{eq:umda}. Function \fnctn{SampleIndividual} is 
used to create the individuals in the initial population (line 3) by sampling the probabilistic model
built. These individuals are decoded (using a guided GRASP algorithm as mentioned before) in order to be 
evaluated (line 4). From there on, the 
algorithm goes into an iterative process until the allowed execution time is exhausted. During each 
iteration, a new population of selected solutions from the current population is built by 
using truncation of the lower half (quality-wise) of the population (line 7), a new model 
is created by using the greedy order of detectors in these solutions (line 8), 
and a new population to be used 
in the next iteration is generated by sampling the new model (lines 9-12).

\section{Experimental Results}
\label{sec:results}
In this section, we analyze the results of different experiments we have carried out regarding the OPSBD problem. First of all, we do a comparison of the performance of the various algorithms introduced in \refSect{sect:Algorithms} on a extensive set of synthetic instances comprising a broad combination of parameters. Next, we do a sensitivity analysis on the performance of the most effective algorithm found in previous comparison, with the aim of understanding how different settings for parameters are related to the difficulty of problem instances. Finally, we evaluate how the best algorithm performs in the case of instances representing real world locations.

Although, as mentioned previously, we have considered a large set of settings for doing these experiments, 
there are some parameters for which we have used values which are good approximations to their typical 
configurations in a real environment and/or are analogous to the values previously used in the literature 
\citep{kaplan05operational,Xiaofeng2007}. In this way, we have assumed for simplicity that the 
probability of the attacker for choosing each of the paths going from one entrance to an objective is constant, 
i.e., $\gamma_{ij} = 1 / \objectives \entrances$, as in \citep{Xiaofeng2007}. We have also contemplated 
that effective neutralization of the attacker is not possible if the distance to the objective is less than 10m 
(here we assume that aborting the attack is plausible if the attacker is at least 10 seconds away from the objective and that the terrorist moves 
at a speed of 1m/s). In the case of the detector's instantaneous detection rate, we used a value of
$\eta=0.06$ and for the probability of effectively neutralizing a detected attacker, we used $\neutral=0.6$. 
Finally, in order to estimate the expected number of casualties for an objective cell $C_{j}, 1\leqslant
j\leqslant\objectives$, we used 
equation [2] in \citep{kaplan05operational} which assumes that the number of fragments after the explosion 
tends to $\infty$ and that these and individuals around the target area follow a spatial Poisson process. We used the same parameters for this equation as those used in \citep{Xiaofeng2007}, except for the population densities near objective cells, which was set as a constant there but we modelled as a random variable in
${\mathcal N(0.4, 0.1)}$ persons / ${\mathrm m}^2$
instead, with the aim of considering more diverse instances (see previous references for full details).

Regarding the hardware platform used, all experiments in this paper were executed on Intel Xeon E7-4870 2.4 GHz processors with 2GB of RAM running under SUSE Linux Enterprise Server 11 operating system.

\subsection{Random instances}

In this section we present results of an experimental comparison on the performance of different proposed heuristics for the OPSBD problem.
We have considered the following algorithms with the parameterization indicated:
\begin{itemize}
	\item The local search \HC{} algorithm presented in \refSect{sect:HC}. This algorithm was run on different random generated instances until the allowed execution time was exhausted.
	\item The \EA{} described in \refSect{sect:EA}. For this algorithm, we used standard parameters 
	and operators: a population size $\mathit{popSize}=100$ individuals, probability of crossover 
	$p_X=0.9$, probability of mutation $p_m=1/\detectors$, binary tournament selection mechanism 
	and replacement of the worst individual. 
	\item The Greedy Randomized Adaptive Procedure introduced in \refSect{sect:GRASP}. In this 
	case, we have considered different settings for parameter $\alpha\in\{0.1,0.25,0.5\}$. We denote 
	these by \GRASP{0.1}, \GRASP{0.25} and \GRASP{0.5} respectively to indicate the value of the 
	parameter in each case. In addition, we 
	have considered a version of this same heuristic that performs a local search on each constructed 
	solution by using the \HC{} algorithm. We denote this latter variant by \GRASPHC{0.1}
	(we have considered this hybrid approach just for $\alpha=0.1$).
	\item The \Greedy{} algorithm originally introduced in \citep{Xiaofeng2007}.
	\item The \umda{} described in \refSect{sect:UMDA}. The parameters for this 
	algorithm were: population size $popSize=100$ individuals and a selected 
	population $pop'$ of 50 individuals. Three different settings for parameter 
	$\alpha$ were used for building the initial probabilistic model, leading to three 
	versions of the algorithm: \UMDA{0.1}, \UMDA{0.25} and \UMDA{0.5}.
\end{itemize}

All algorithms have been run on random instances constructed using the following 
combinations of parameters:
\begin{itemize}
	\item Maps with dimensions of $32 \times 32$ and $64 \times 64$ cells.
	\item Different side lengths for cells comprising the maps: $\cellSz \in \{5,10,20\}$ meters.
	\item Different number on entrances on each border of the map in $\{2,3,4\}$ leading to  $\entrances \in \{8,12,16\}$.
	\item Different number of objective cells on the map: $\objectives \in \{2,4,6,8\}$.
	\item A percentage of blocked cells on the map of 5\%.
	\item Different number of detectors to be placed on the map: $\detectors \in \{6,8,10\}$.
	\item A detection radius $\radius=20$ meters.		
\end{itemize}

For each combination of these parameters, 25 random instances were considered, thus this benchmark comprises a total 
of 5,400 
different problem instances. A maximum execution time of 5 seconds was allowed for each algorithm in the case of $32 \times 32$ instances, whereas the execution time allowed for $64 \times 64$ instances was 10 seconds. 
Since the cache data structure described in \refSect{sect:cache} has to be precomputed by all compared algorithms, its computation time is not included in these time limits.

\begin{figure}[!t]
\subfloat[][]{\centerline{\includegraphics[angle=0,scale=0.37]{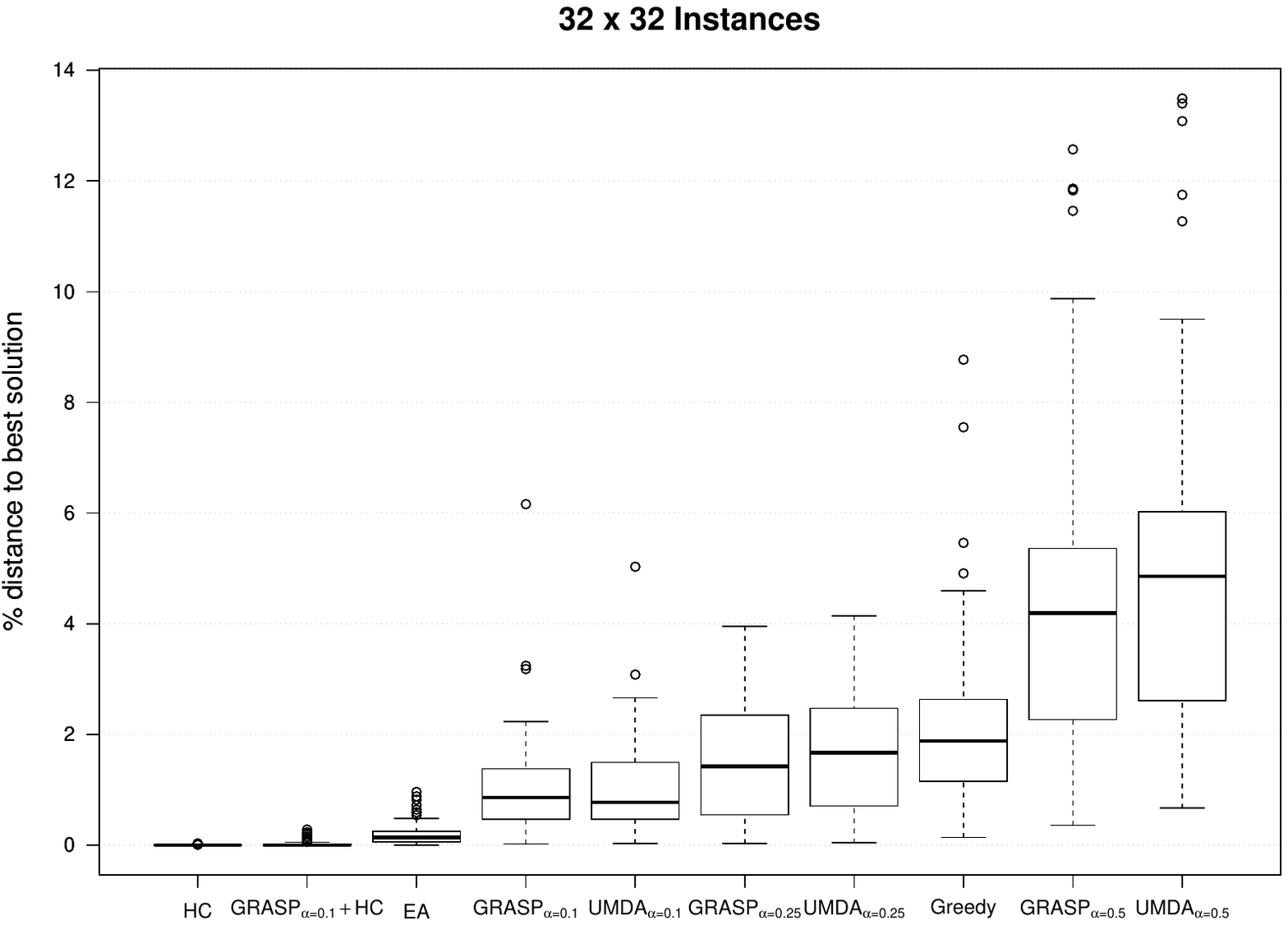}}}

\subfloat[][]{\centerline{\includegraphics[angle=0,scale=0.37]{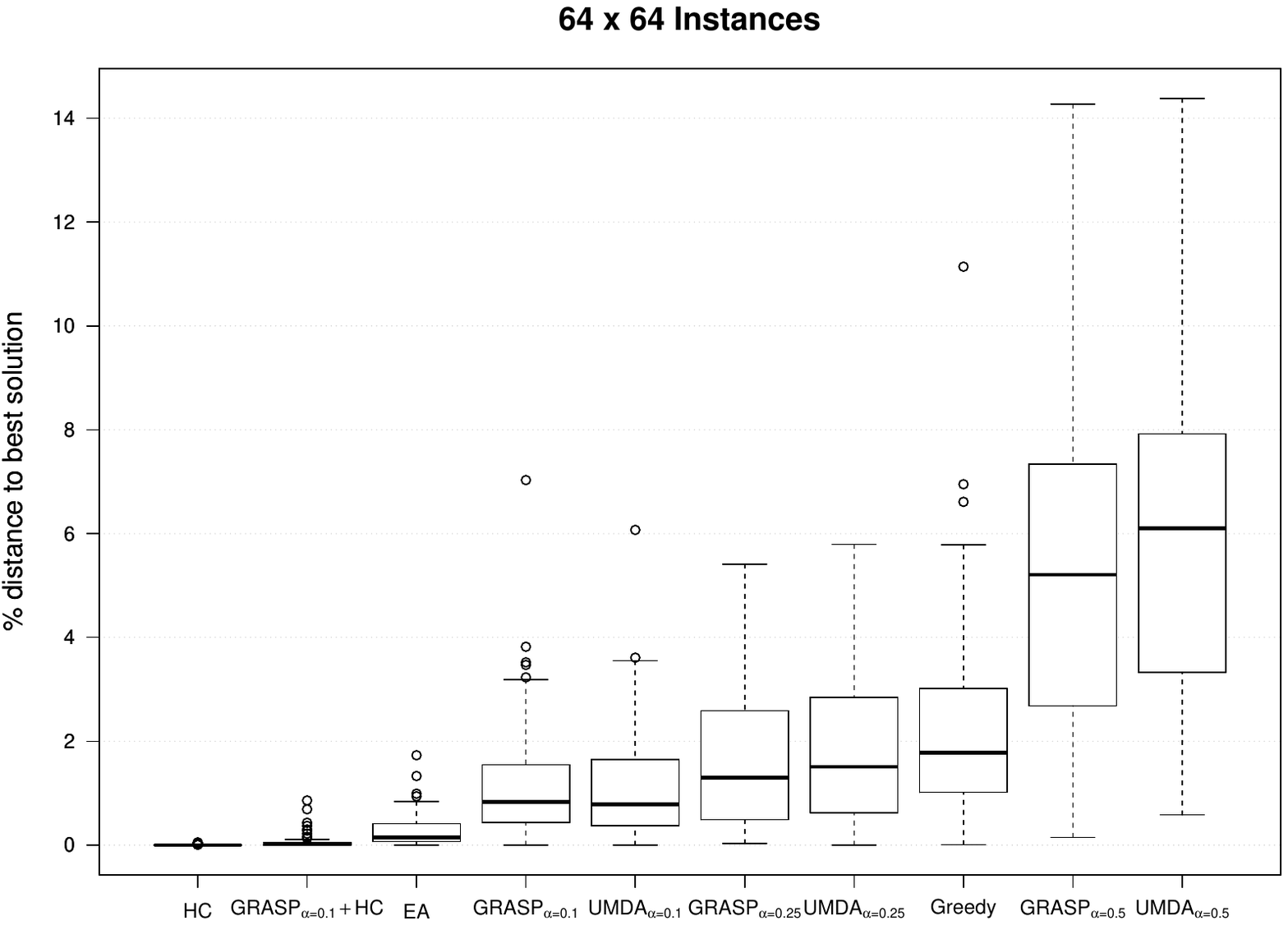}}}

		\caption{Relative distances to best solutions of results provided by different algorithms on (a) $32 \times 32$ instances and (b) $64 \times 64$ instances.}
		\label{fig:allAlgs:32-64} 
\end{figure}

In order to measure performance in a comparable way across the huge collection of problem 
instances considered, we have recorded the best results found by any algorithm on each instance, 
and used these to calculate relative deviations to such best known solutions. The results are 
summarized in \refFig{fig:allAlgs:32-64} showing box plots of the distribution of relative deviations 
to best known solutions for different algorithms, separately for the $32 \times 32$ and $64 \times 64$
instances. It can be seen that deviations are very close to 0 for HC, which means that this algorithm 
was able to provide best 
known
solutions for most of the instances. The EA comes third, with distances that are also quite close to 0. 
\GRASP{0.1} and \GRASP{0.25} are the next two better algorithms, but their performance degrades 
clearly when compared to HC and EA. The value of upper whisker for the distribution corresponding 
to \Greedy{} on $32 \times 32$ instances is of a 4.59\% relative distance to the optimum, 
whereas the same value for HC  is 0\% and for EA is 0.48\%. The difference is of a 5.78\% in the 
case of $64 \times 64$ instances. This is an indications of the extent of improvements achieved
by both heuristics over \Greedy{}.

Subsequently, a statistical test was used in order to analyze whether differences 
among the different techniques are statistically significant. More precisely, 
the Friedman Test (FT) \citep{Friedman1937Use,friedman1940}, a non-parametric 
test based on rankings, has been used for this purpose. If, as a result of 
this test, the null hypothesis stating equality of
rankings between the different techniques is rejected, we proceed to post-hoc procedures in 
order to detect statistical differences among different pairs of algorithms. For this purpose, 
Shaffer's static procedure \citep{Shaffer86} with a standard significance level of $\alpha=0.05$ 
has been used. All these analysis were carried out using the software package provided by 
the Soft Computing and Intelligent Information Systems group at University of Granada \citep{garcia2008extension}.

\begin{table}[!t]
	\caption{Mean ranks of all algorithms on $32 \times 32$ and $64 \times 64$ cells instances.}
	\label{friedman:allAlgs:3264}       
	\begin{tabular}{llr}
		\noalign{\smallskip}
		\hline\noalign{\smallskip}
		position & algorithm &ranking\\
		\noalign{\smallskip}\hline\noalign{\smallskip}
		1\textsuperscript{st}  & HC  & 1.28 \\
		2\textsuperscript{nd}  & \GRASPHC{0.1}  & 1.83 \\
		3\textsuperscript{rd}  & EA  & 3.17 \\
		4\textsuperscript{th}  & \GRASP{0.1} & 4.82 \\
		5\textsuperscript{th}  & \UMDA{0.1} & 4.93 \\
		6\textsuperscript{th}  & \GRASP{0.25}  & 6.08 \\
		7\textsuperscript{th}  & \UMDA{0.25} & 6.75 \\
		8\textsuperscript{th}  & \Greedy{} & 7.39 \\
	    9\textsuperscript{th}  & \GRASP{0.5}  & 8.95 \\
   	   10\textsuperscript{th}  & \UMDA{0.5} & 9.80 \\
		\noalign{\smallskip}\hline
	\end{tabular}
\end{table}

Firstly, the mean rank of all ten algorithms in the benchmark considered is computed and shown in \refTable{friedman:allAlgs:3264}. As can be seen, the order of performance from best to worst is: HC, followed by EA, the two variants of GRASP$|$UMDA using smallest values of $\alpha$, the \Greedy{} algorithm and finally GRASP$|$\UMDA{0.5}.
To determine the significance of these rank differences, the FT is performed 
and results in $\chi^2_F=1726.18$, much 
larger than the critical value 16.92 for $\alpha=0.05$ (according to $\chi^2$ with 9 degrees of freedom).
The $p$-value calculated by this test is actually $p=0$ which provides strong evidence for rejecting 
equality of rankings. Shaffer's post-hoc procedure, at the same standard $\alpha=0.05$ level, found
statistically significant differences between all pairs of algorithms, except for 
\GRASP{0.1}  vs \UMDA{0.1},
\GRASP{0.25}  vs \UMDA{0.25},
\GRASP{0.5}  vs \UMDA{0.5},
\UMDA{0.25} vs \Greedy,
and HC vs \GRASPHC{0.1}. It can thus be concluded that using the \GRASP{0.1} component 
for initializing solutions for the HC algorithm is not advantageous with respect to using random 
initialization for this same purpose. Moreover, using the UMDA in order to guide the search of 
GRASP does not provide benefits with respect to taking random decisions, as it is done 
in the pure GRASP algorithm. 
From a more general point of view, all the algorithms proposed in this paper have a better performance than the \Greedy{} algorithm from \citep{Xiaofeng2007}, except \GRASP{0.5} and \UMDA{0.5} (statistically worse) and \UMDA{0.25} (better than \Greedy{} but not significantly so). It can be also seen that the performance of GRASP algorithm improves with smaller values of $\alpha$, but this algorithm performs worse than HC and EA for this problem.

\begin{figure}[!t]
	\begin{center}
		\includegraphics[angle=0,width=.47\textwidth]{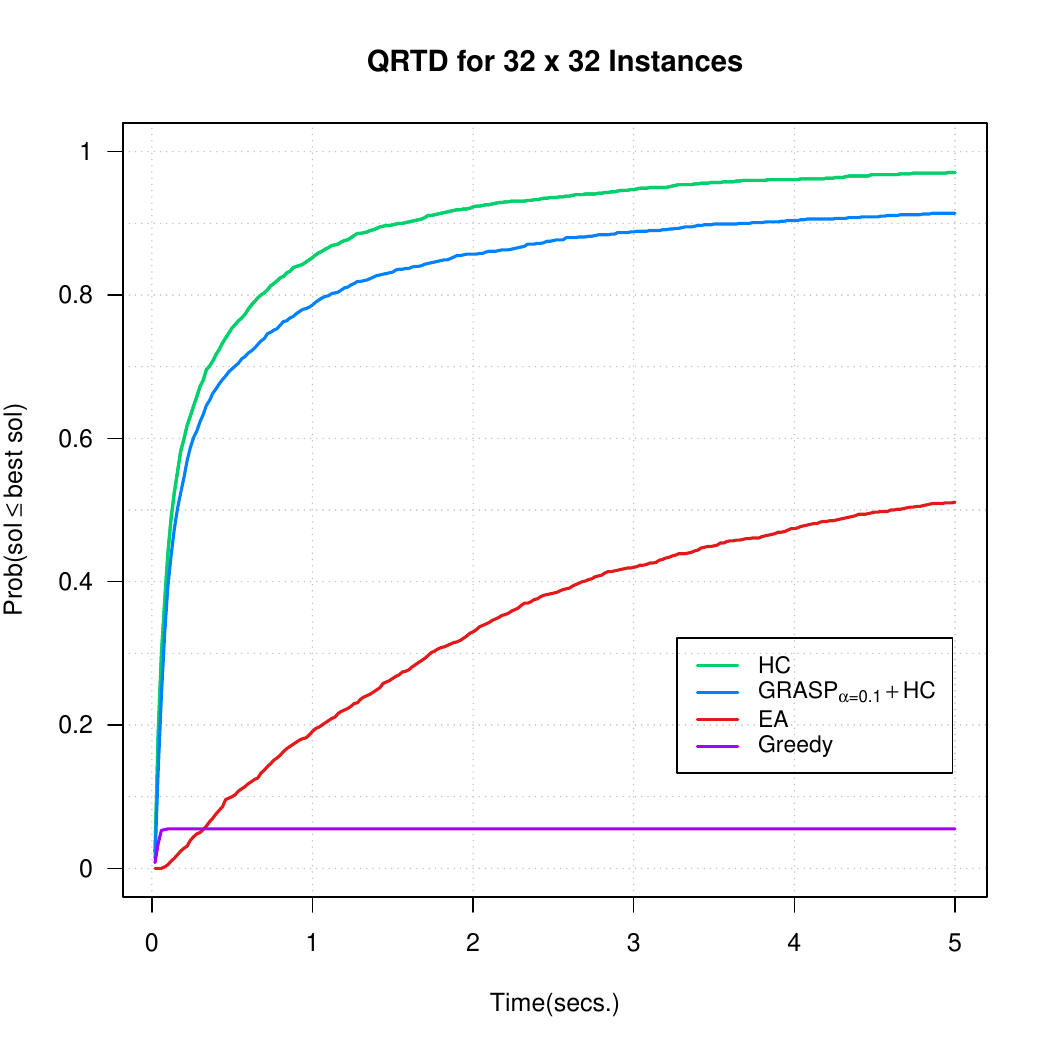}~
		\includegraphics[angle=0,width=.47\textwidth]{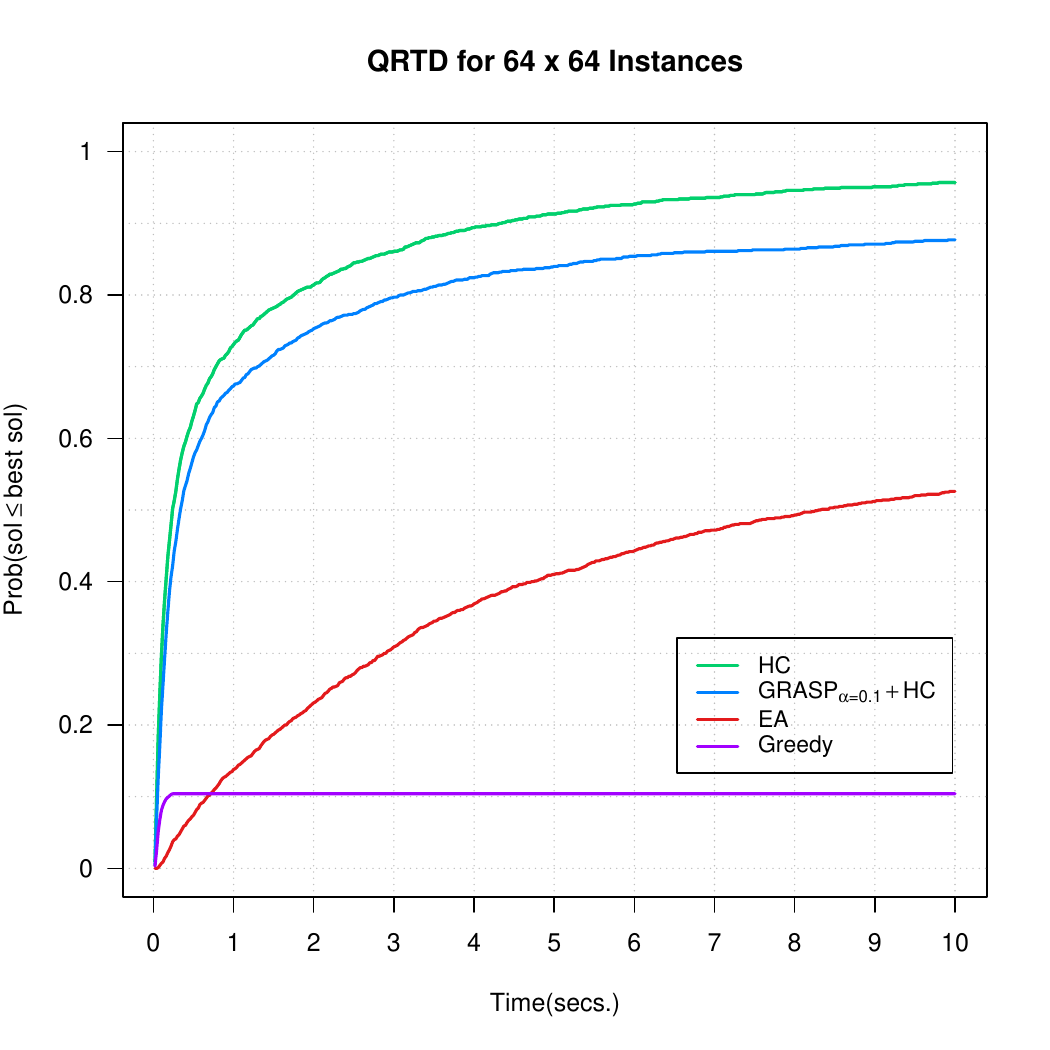}	
		\caption{Qualified runtime distributions (QRTD) for best performing algorithms (\Greedy{} also included for comparison purposes) on $32 \times 32$ (left) and $64 \times 64$ (right) instances. Curves show the probability of reaching best known solutions along time for different algorithms.}
		\label{fig:QRTD} 
	\end{center}	
\end{figure}

The behavior as anytime 
algorithms of both the best performing heuristics and \Greedy{} is analyzed in 
\refFig{fig:QRTD}. This figure shows Qualified Runtime 
Distributions (QRTD) for the best three algorithms, namely 
HC, \GRASPHC{0.1} and EA on $32 \times 32$ and $64 \times 64$ instances. 
Such QRTDs show the probabilities of reaching best known solutions as a 
function of execution time. It can be seen that the algorithm that progresses 
faster towards high quality solutions is HC. This algorithm is able to find best 
solutions for 90\% of the $32 \times 32$ instances in about 1.5 seconds. For 
$64 \times 64$ instances, the same rate of success is achieved after 4.0 seconds 
of execution. Taking into account the context in which the OPSBD problem should 
be solved, these seem to be sensible time requirements, amenable for
fast-response adjustment of detector locations. From a more detailed point
of view, the previous analysis showed 
that HC and \GRASPHC{0.1} were roughly equivalent with respect to the quality of 
solutions they provide, but QRTDs show that \GRASPHC{0.1} is however slightly 
slower than HC. As \GRASPHC{0.1} is a more complicated algorithm to implement 
and also does not provide advantages over HC, we can conclude that using the latter 
one is a better alternative to solve this problem. Regarding the EA, QRTDs also show 
that it progresses at a slower rate than HC, as its curve is always below the one for 
HC. Finally, although \Greedy{} is not able to progress towards 
best solutions (it is a deteministic constructive algorithm) its inclusion in the QRTDs allows
comparing its performance to the remaining algorithms on a time-quality basis. 
Notice that compared to the EA, it provides a better solution initially, but the
EA can outperform \Greedy{} in a matter of tenths of a second. This is exactly the 
expected behavior for such a kind of algorithm, as Greedy local decisions can be 
made in a very fast way but they are expected to be suboptimal in general. 

\begin{figure}
	\begin{center}
		\includegraphics[angle=0,scale=0.37]{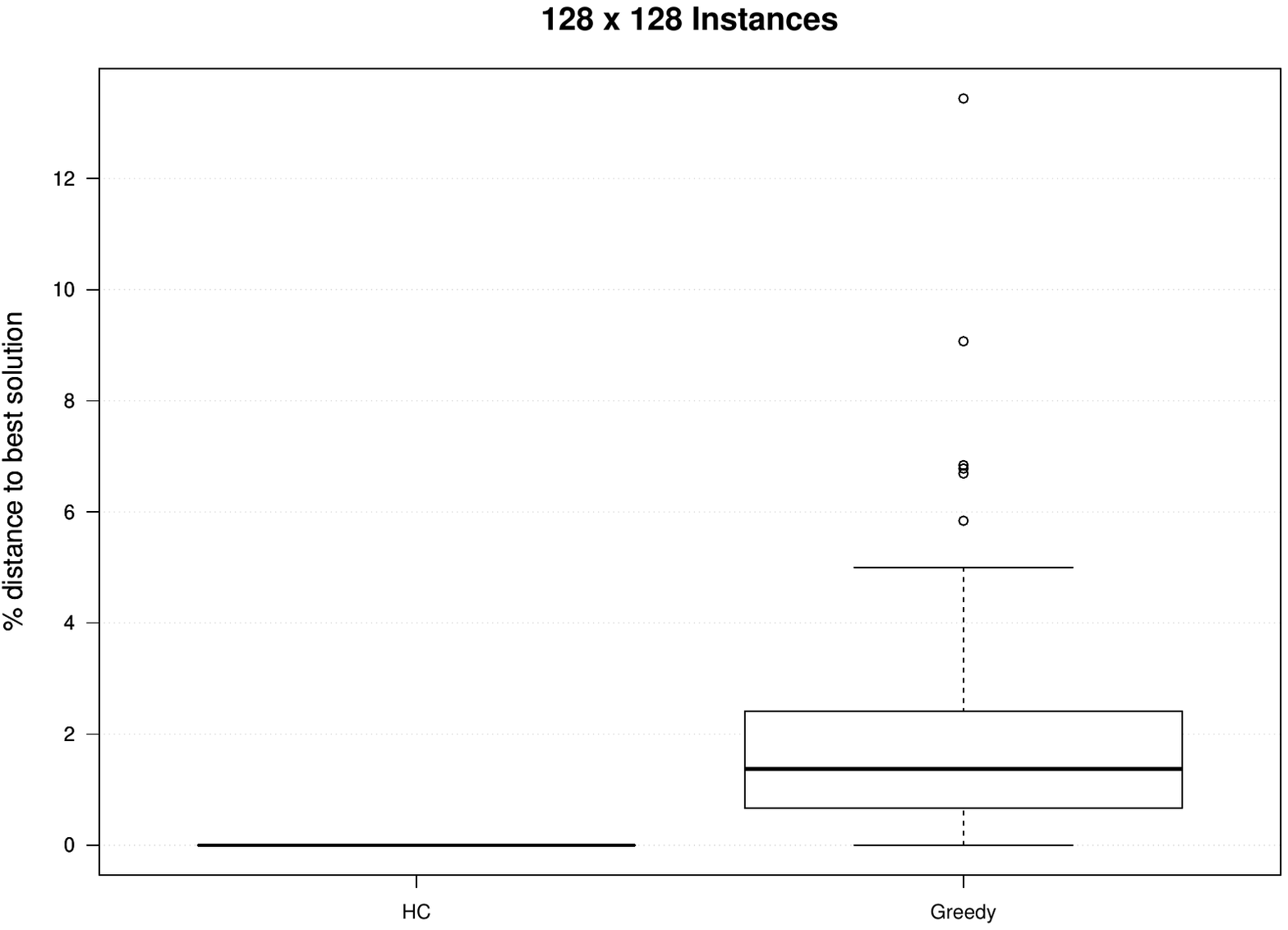}
		\caption{Relative distances to best known solutions of results provided by HC and \Greedy{} on $128 \times 128$ instances.}
		\label{fig:allAlgs:128} 
	\end{center}	
\end{figure}

Finally, we compared the best performing algorithm (HC) with the previous proposal in the literature (\Greedy) on larger random instances. In this case, maps consisted of $128 \times 128$ grids of cells and, for the rest of parameters, we considered the same combinations as those we used previously for $32 \times 32$ and  $64 \times 64$ instances. Maximum allowed execution times were 10 seconds per instance in this case. A Wilconxon signed ranked statistical test rejects the null hypothesis stating equality of rankings between the two algorithms ($p$-value=0). Boxplots for relative distances to best known results are shown in \refFig{fig:allAlgs:128}. It can be seen that HC algorithm also provides consistently best results for these instances, and that \Greedy{} is not able to provide solutions with same quality. The value of upper whisker for the distribution corresponding to these instances of \Greedy{} is at a 5\% relative distance to the optimum, whereas the same value for HC algorithm is 0\%.

\subsection{Sensitivity Analysis}
\label{sect:Sensitivity:Analysis}

In this section, we perform a sensitivity analysis on the performance of the \HC{} algorithm
(\refSect{sect:HC}) with respect to \Greedy{} (\refSect{sect:GREEDY}) with 
the aim of studying which parameter settings of the problem lead to a larger performance gap
among these two techniques.  The rationale for this procedure is that if the relative improvement 
of the solution provided by \HC{} over \Greedy{} increases for some parameter 
settings, it can be considered that those instances turn out to be harder-to-solve for the latter. 
The different parameters and settings that have been analyzed are a superset of those considered
in the previous section, namely:
\begin{itemize}
\item Maps with dimensions of $32 \times 32$ and $64 \times 64$ cells.
\item Different side lengths for cells comprising the maps: $\cellSz \in \{5,10,20\}$ meters.
\item Different number on entrances on each border of the map in $\{2,3,4\}$ leading to  $\entrances \in \{8,12,16\}$.
\item Different number of objective cells on the map: $\objectives \in \{2,4,6,8\}$.
\item Different percentages of blocked cells on the map: $\blocked \in \{2.5\%, 5\%, 10\%\}$.
\item Different number of detectors to be placed on the map: $\detectors \in \{2,4,6,8,10\}$.
\item Different values of the detection radius: $ \radius \in \{10,20,40\}$ meters.		
\end{itemize}

All possible combinations of these parameter settings lead to 3,240 different kinds of 
instances, and for each of them, we have considered 25 random instances, so that our 
analysis comprises a total of 81,000 different instances. In the case of $32 \times 32$ 
cell maps, the maximum execution time allowed for the algorithms on each instances was 
5 seconds whereas in the case of $64 \times 64$ maps, this time was extended to 10 seconds.

\begin{figure}[!t]
	\begin{center}
		\includegraphics[angle=0,scale=0.37]{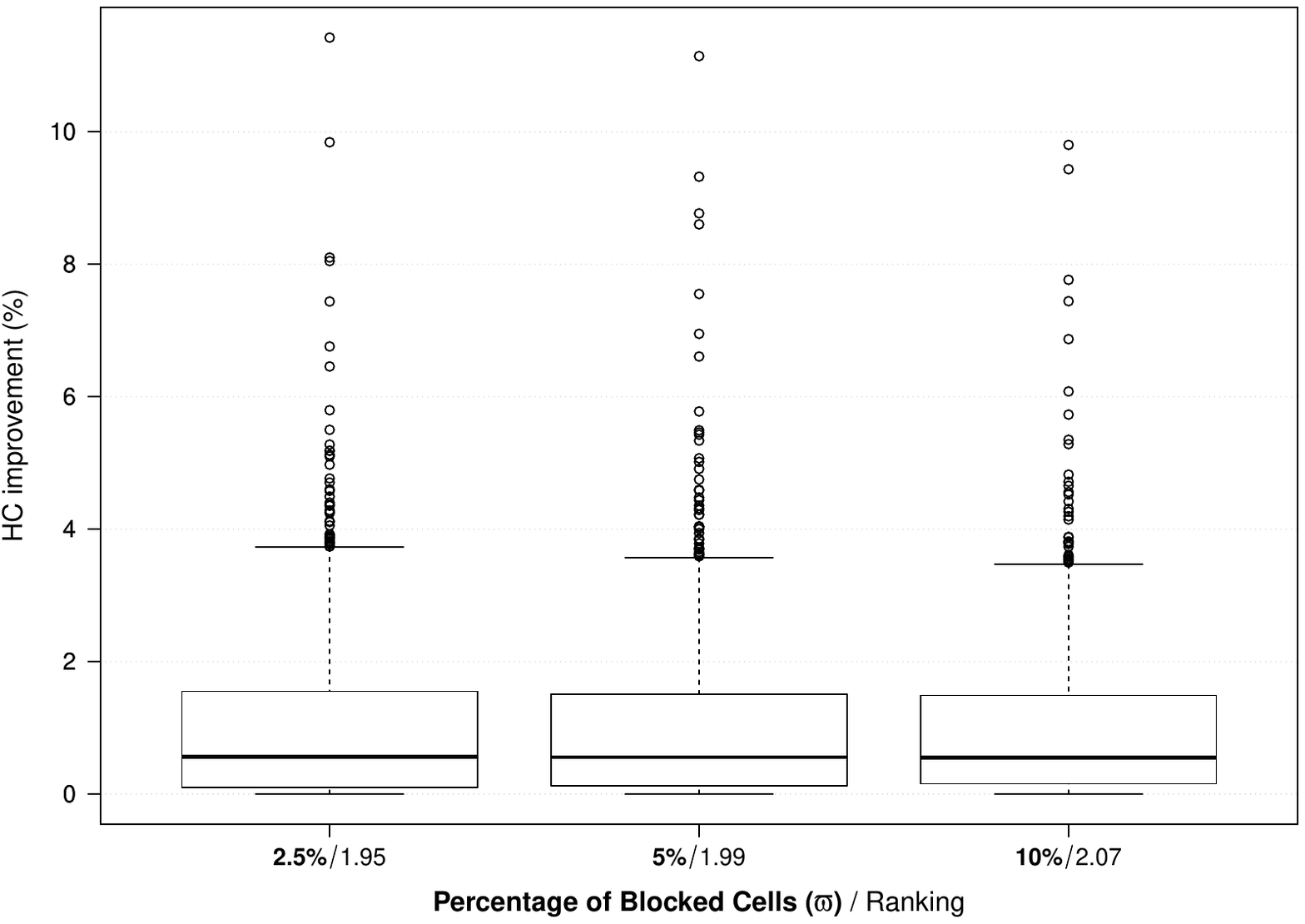}
		\caption{Relative improvements of HC over \Greedy{}  for different percentages of blocked cells (\blocked) in the map. Settings are ordered (left to right) from more difficult to easier ones. Rankings are shown at the right side of labels. In order to interpret this figure and subsequent ones in this subsection, notice that the boxplots depict the distribution of the aggregated data and show the range of variation in each case, whereas the rankings are computed by aligning the results obtained by each algorithm on the instances generated by each parameter combination and are used by the statistical tests. }
		\label{fig:friedman:blocked} 
	\end{center}	
\end{figure}

\subsubsection{Percentage of Blocked Cells}
We firstly analyzed differences between maps with varying percentages of blocked cells 
($\blocked \in \{2.5\%, 5\%, 10\%\}$). The ranking is shown in
\refFig{fig:friedman:blocked}. The value of the Friedman statistic is $\chi^2_F=7.59$ that 
corresponds to a $p$-value = $0.02$  (the critical value is in this case is distributed according 
to $\chi^2$ with 2 degrees of freedom, yielding $5.99$). 
This provides evidence for rejecting the null hypothesis 
that states equality of
rankings between the different settings for \blocked{}.
The order in \refFig{fig:friedman:blocked} goes from more difficult instances to  easier ones, 
so that the first position corresponds to the setting for which the \HC{} algorithm provides 
the largest improvement. Post-hoc procedures only show statistical significant differences, at 
the standard significance level of $\alpha=0.05$, between 
configurations for \blocked{} parameter of 2.5\% vs 10\%. Thus, it can be seen that a lower 
percentage of blocked cells leads to maps that are harder to solve for \Greedy, 
but also that there are no differences between the intermediate setting for this parameter and 
the smaller or larger ones.
We interpret this result in terms of the broader dispersion of shortest paths in instances 
with fewer obstacles, which makes more difficult to find major cross points and hence
the goodness of solutions relies on the distribution of detectors with a more global perspective,
as opposed to covering a couple of critical points and use the remaining detectors for 
fine tuning.

\subsubsection{Side Length of Cells}
\begin{figure}[!t]
	\begin{center}
		\includegraphics[angle=0,scale=0.37]{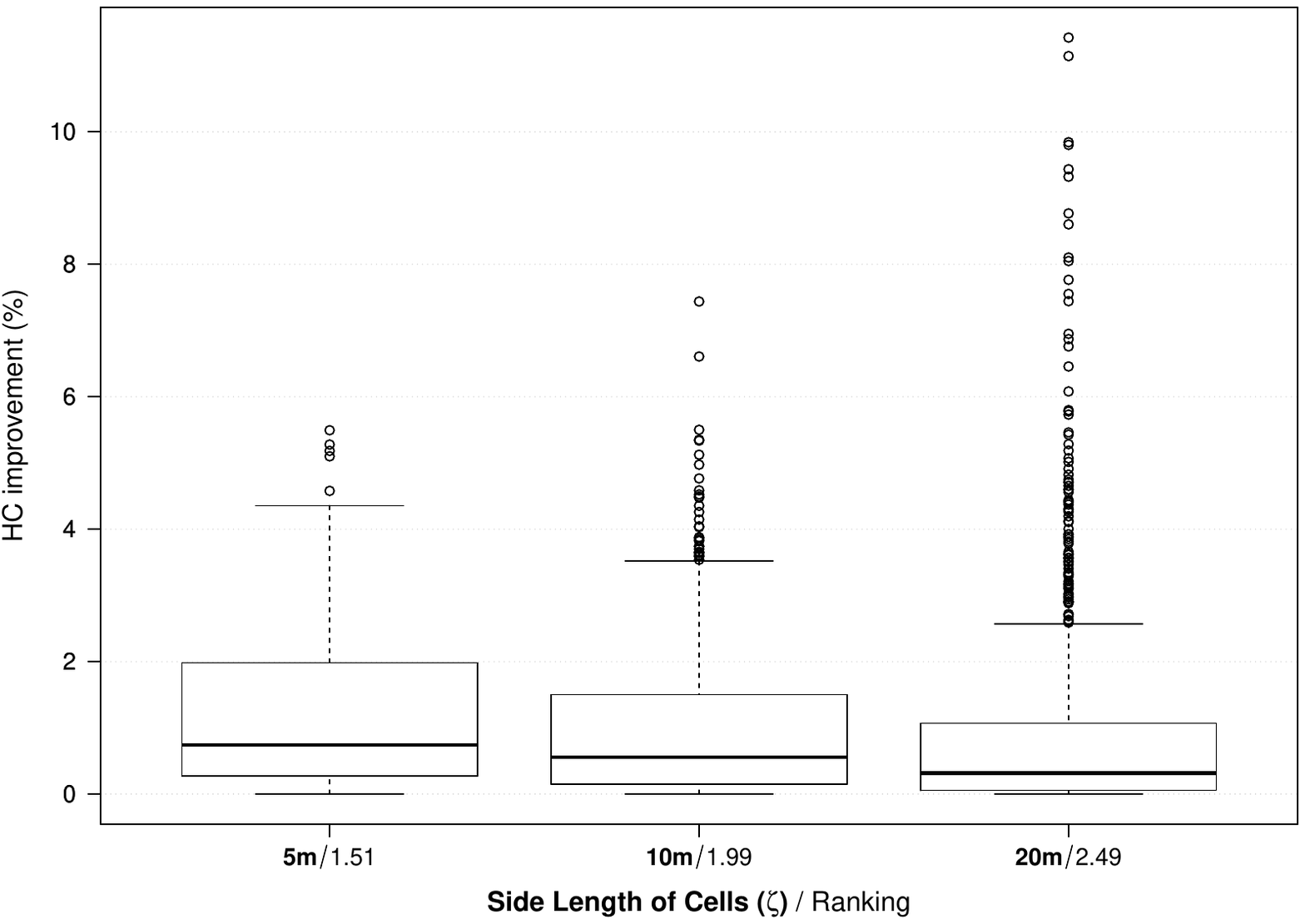}
		\caption{Relative improvements of HC over \Greedy{}  for different side lengths (\cellSz) of cells in the grid. Settings are ordered (left to right) from more difficult to easier ones. Rankings are shown at the right side of labels.}
		\label{fig:friedman:cellSz} 
	\end{center}	
\end{figure}

Regarding side lengths of cells (\cellSz) in the grid, the ranking is shown in
\refFig{fig:friedman:cellSz}. We obtain $\chi^2_F=518.25$, corresponding to a
$p$-value = $2.34\cdot10^{-10}$ (the critical value is here distributed
again according to $\chi^2$ with 2 degrees of freedom, yielding $5.99$). 
Thus, there is strong evidence for rejecting the null hypothesis that states equality of
rankings between the different settings for \cellSz{}. In this case, post-hoc procedures found statistically significant differences (at 
the standard significance level of $\alpha=0.05$) between all possible settings for this parameter. 
It can be seen that the difficulty of instances increases as cell sizes decrease. One possible 
explanation for this phenomenon may be that having a more fine-grained map allows for a more 
precise placement of detectors in the map. This seems to be better exploited by \HC, that has 
more choices for fine-tuning the solution from a global perspective (as opposed to the local
approach of \Greedy).

\begin{figure}[!t]
	\begin{center}
		\includegraphics[angle=0,scale=0.37]{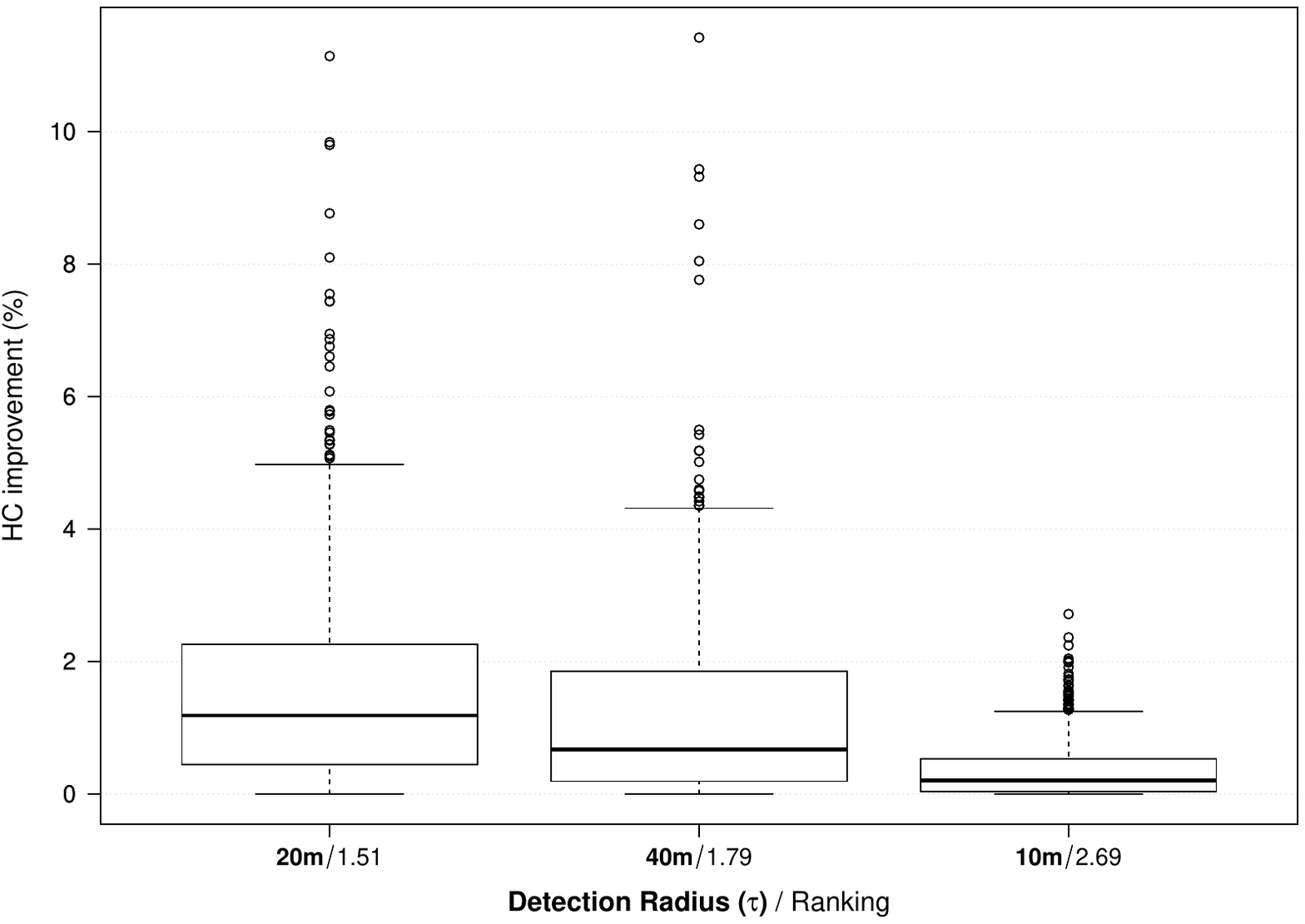}
		\caption{Relative improvements of HC over \Greedy{}  for different values of the detection radius (\radius). Settings are ordered (left to right) from more difficult to easier ones. Rankings are shown at the right side of labels.}
		\label{fig:friedman:radiuses} 
	\end{center}	
\end{figure}

\subsubsection{Detection Radius}
The detectors placed in the map may have a different radius of effectiveness. Here we 
consider settings for this parameter in $\radius \in \{10{\rm m},20 {\rm m},40 {\rm m}\}$. 
The resulting ranking is shown in \refFig{fig:friedman:radiuses}. We obtain $\chi^2_F=825.17$,
corresponding to a $p$-value = $2.9\cdot10^{-10}$ (the critical value is once again distributed
according to $\chi^2$ with 2 degrees of freedom). Hence, the null hypothesis stating equality of
rankings between the different settings for this parameter should be rejected. In the case of this parameter, post-hoc procedures found statistically significant differences (at the standard significance level of $\alpha=0.05$) between all pairs of settings. 
It must be noted that for this parameter harder instances are those with an intermediate radius of detection. One interpretation of these results may be that a smaller radius leads to less interactions between the detectors and paths followed by terrorists, whereas a very large radius allows for one detector to cover a great extent of the map, thus simplifying the difficulty of the problem. It is the intermediate radius $\radius =20$m that leads to harder instances in this case as there is a good degree of interaction of each detector with different paths and, at the same time, the radius is small enough for a precise placement of the detector to turn out to be crucial. Of course, this effect 
also depends on the number of detectors to be placed since, if this number is large, a suboptimal placement of some detectors may be to some extent compensated by other detectors covering the same paths. In this case,
the smaller the detection radius, the more influential the location of the detectors would become.

\subsubsection{Number of Detectors}

\begin{figure}[!t]
	\begin{center}
		\includegraphics[angle=0,scale=0.37]{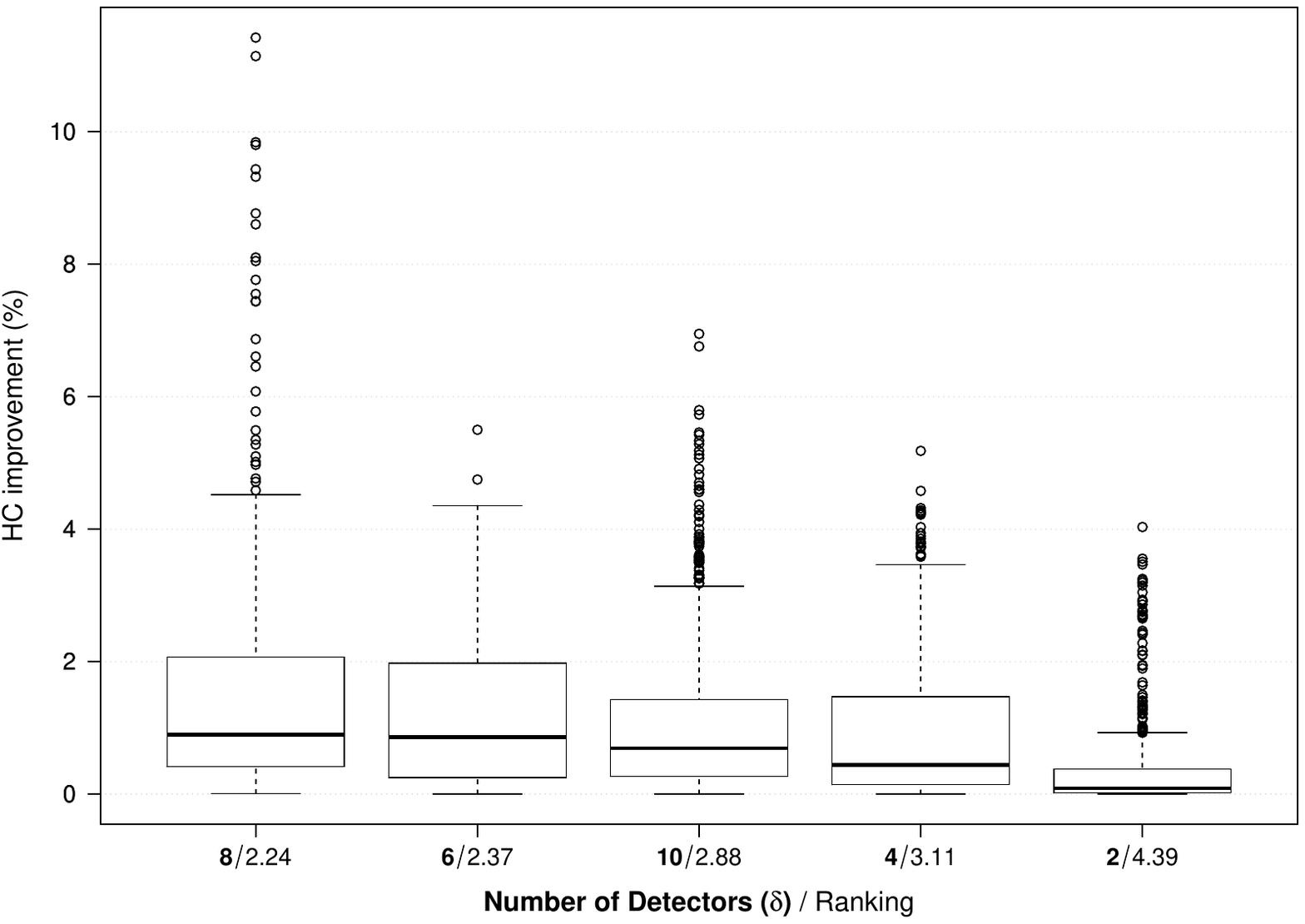}
		\caption{Relative improvements of HC over \Greedy{}  for different number of detectors (\detectors) placed on the map. Settings are ordered (left to right) from more difficult to easier ones. Rankings are shown at the right side of labels.}
		\label{fig:friedman:detectors} 
	\end{center}	
\end{figure}

We have considered different number of detectors (\detectors) to be placed on the map. Rankings are shown in \refFig{fig:friedman:detectors}. The associated Friedman statistic is $\chi^2_F=763.50$ which implies a $p$-value = $2.41\cdot10^{-10}$ (the critical value is in this case distributed
according to $\chi^2$ with 4 degrees of freedom, yielding $9.49$) and  
thus there are significant differences for different settings of this parameter. Post-hoc procedures found statistically significant differences (at 
the standard significance level of $\alpha=0.05$) between all pairs of settings except for 6 vs. 8. Results show that instances with a smaller number of detectors ($\detectors \in\{2,4\}$) are the easiest ones, but also that instances with the maximum number of detectors (\detectors=10) are easier than those with 8 and 6 detectors. One interpretation of these results is that the search space for a small number of detectors is relatively easy to explore for different algorithms, but also that having a higher number of detectors allows for placing them in a less precise way without affecting in a significant way the quality of the solution. In other words, instances with an intermediate number of detectors lead to a large enough search space which raises the difficulty of the
problem. At the same time, a precise placement of the limited number of available detectors is required in order
to provide high-quality solutions for these instances.

\subsubsection{Number of Entrances}
\label{sec:stats:entrances}

\begin{figure}[!t]
	\begin{center}
		\includegraphics[angle=0,scale=0.37]{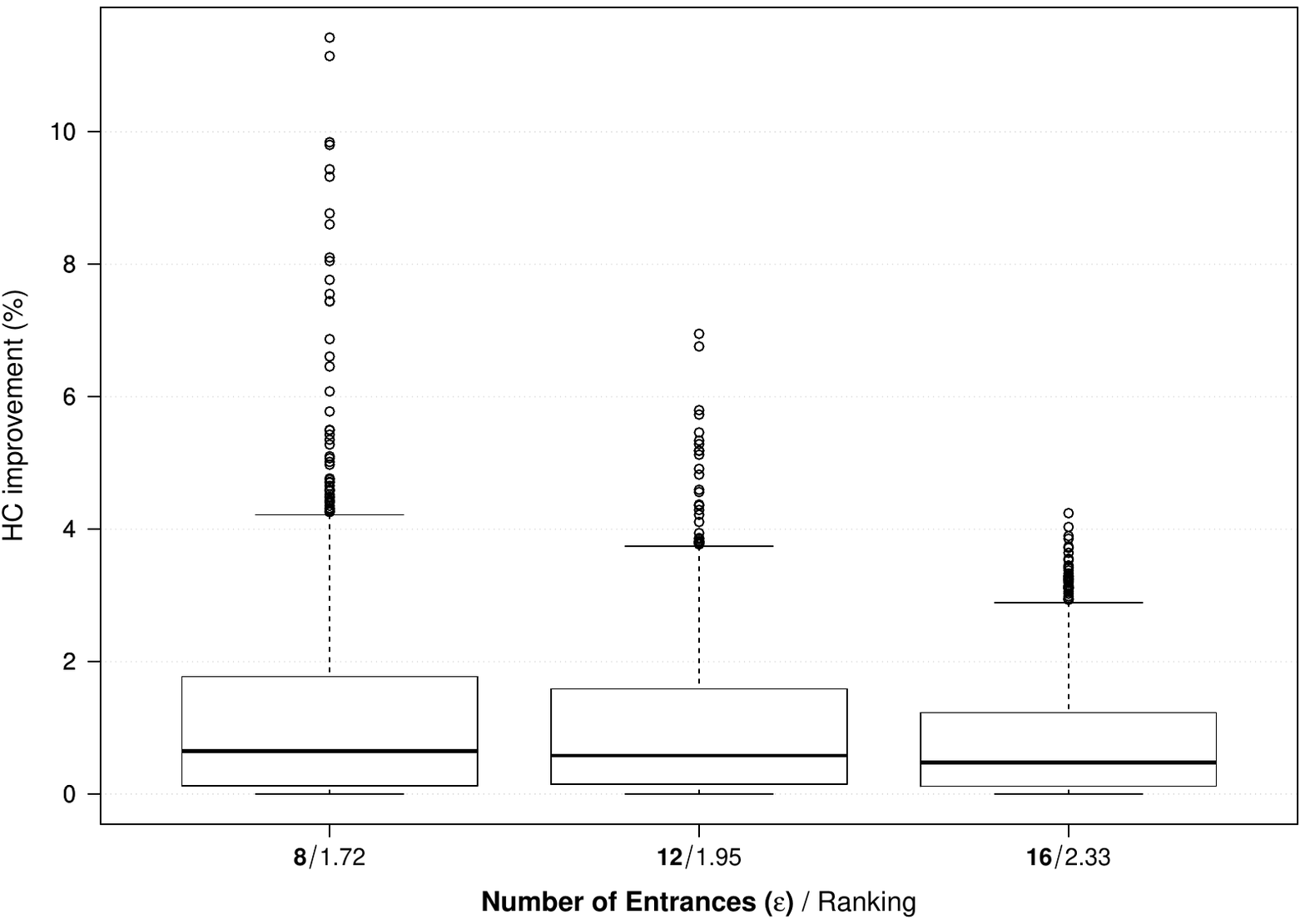}
		\caption{Relative improvements of HC over \Greedy{}  for different number of entrances (\entrances) on the map. Settings are ordered (left to right) from more difficult to easier ones. Rankings are shown at the right side of labels.}
		\label{fig:friedman:entrances} 
	\end{center}	
\end{figure}

Regarding the number of entrances on the map (\entrances), we have considered 2, 3 and 4 entrances on each side of the map, which leads to the following settings:  $\entrances \in \{8,12,16\}$. Rankings are shown in \refFig{fig:friedman:entrances}. The Friedman statistic is $\chi^2_F=202.47$ and the corresponding $p$-value 
is $9.97\cdot10^{-11}$
(the critical value is in this case distributed
according to $\chi^2$ with 2 degrees of freedom, yielding $5.99$). 
Post-hoc procedures found differences (at 
the standard significance level of $\alpha=0.05$) between all pairs of settings for this parameter. In this case, the larger the number of entrances, the easier the resulting problem instances are. This result can be interpreted in terms
of the number of paths (which is proportional to the number of entrances). Recall that the probability of targeting a certain objective is distributed across the different paths emanating from each entrance. Therefore, the larger the number of entrances, the smaller the weight of individual paths. This means that adjusting the precise location of a detector will yield a smaller variation of the objective function as a result of the the different coverage of the numerous existing paths. Of course, we cannot take this argument to the opposite limit because if the number of entrances was very low it would be easy to cover the paths by any appropriate heuristic.

\subsubsection{Number of Objectives}

\begin{figure}[!t]
	\begin{center}
		\includegraphics[angle=0,scale=0.37]{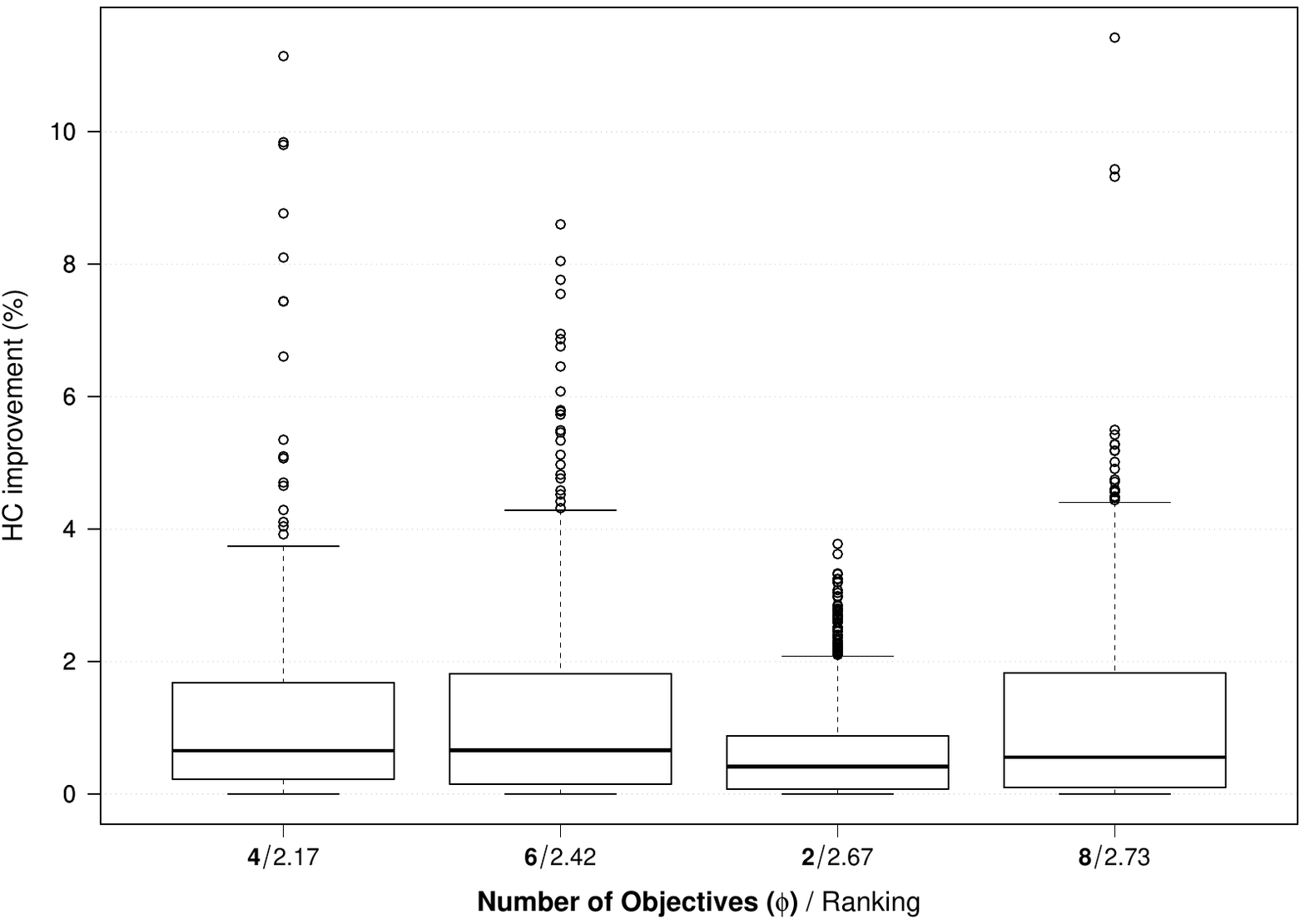}
		\caption{Relative improvements of HC over \Greedy{}  for different number of objectives (\objectives)  on the map. Settings are ordered (left to right) from more difficult to easier ones. Rankings are shown at the right side of labels.}
		\label{fig:friedman:objectives} 
	\end{center}	
\end{figure}

Here we consider maps with different numbers of objective cells ($\objectives \in \{2,4,6,8\}$). Rankings
are shown in \refFig{fig:friedman:objectives}. 
The Friedman statistic is $\chi^2_F=94.69$, implying a $p$-value = $5,74\cdot10^{-11}$
(the critical value is in this case distributed
according to $\chi^2$ with 3 degrees of freedom, yielding $7.81	$). 
Post-hoc procedures in this case show that there are not statistically significant differences (at 
the standard significance level of $\alpha=0.05$) between maps with 2 and 8 objectives. In this case, instances with the lowest ($\objectives=2$) and highest ($\objectives=8$) number of objectives are thus indistinguishable, and both lead to easiest-to-solve instances. For intermediate settings of this parameter, instances with $\objectives=4$ are harder-to-solve than those with $\objectives=6$. 
These results are partly explained in accordance to the argument laid down in \refSect{sec:stats:entrances} for the number of entrances, since the number of paths is also proportional to the number of objectives. In this sense, the lower limit ($\objectives = 2$) correspond to instances in which there can be little variation of quality in solutions because most paths leading to objectives can be adequately covered. Similarly, in the upper limit ($\objectives = 8$) the detectors need to cover much ground and as a result adjusting the location of a detector placed by \Greedy{} can increase the coverage of some paths at the expense of others, resulting in the smallest window of improvements.

\subsection{Real Instances}

\begin{figure}[!t]
\subfloat[][\label{fig:mapsReal:map1}]{\adjustbox{trim=0 {0\height} 0 {0\height},clip}{\includegraphics[scale=0.8,angle=0,width=.3\textwidth]{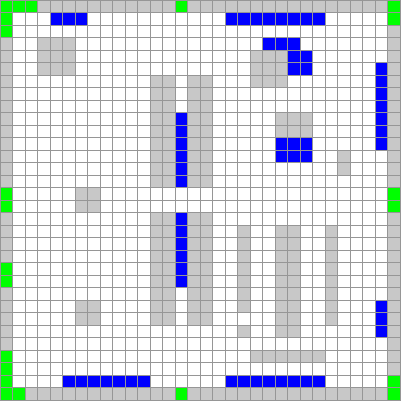}}}
~\hspace{.025\textwidth}~
\subfloat[][\label{fig:mapsReal:map2}]{\adjustbox{trim=0 {0\height} 0 {0\height},clip}{\includegraphics[scale=0.8,angle=0,width=.3\textwidth]{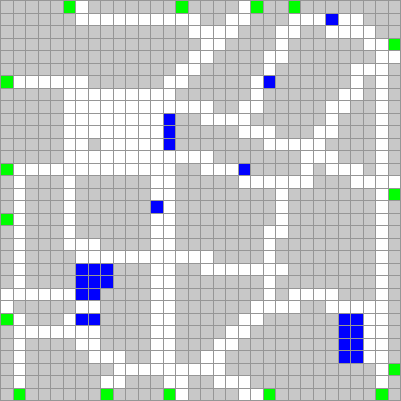}}}
~\hspace{.025\textwidth}~
\subfloat[][\label{fig:mapsReal:map3}]{\adjustbox{trim=0 {0\height} 0 {0\height},clip}{\includegraphics[scale=1,angle=0,width=.3\textwidth]{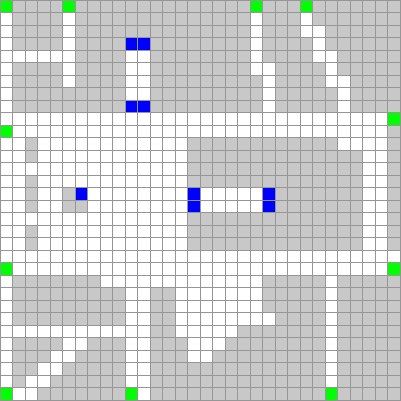}}}

\caption{Realistic maps considered. 
(a) A large plaza in the city center. (b) An old-town urban area with curvy streets and small plazas. (c) A mixed scenario including both a large plaza and some adjacent streets. 
\label{fig:mapsReal} }

\end{figure}

Finally, we have deployed \HC\ and \Greedy{} on some 
more realistic instances resembling the features exhibited by
urban areas that might be subject to a terrorist attack. These are
depicted in \refFigs{fig:mapsReal:map1}{fig:mapsReal:map3} and comprise
three different scenarios: (1) a large square, (2) an old-town area and (3)
a mixture of the previous two -- see \refTable{tab:scenarios}. 
All of them have been constructed based on real-world locations.
Objectives in these scenarios correspond to typical crowded places
such as cafe terraces, small flea-markets or bazaars, or monuments
with great touristic attractive and other iconic landmarks that gather 
people around them. All scenarios are represented on a $32\times32$ grid.
These instances will be used to test the comparative performance of
both algorithms (\HC\ and \Greedy) in a more pragmatic context, 
identifying the circumstances under which there is a larger quality gap
between them and how different the solutions provided by either
algorithm are from a structural point of view.

\begin{table}[!t]
\caption{Description of the three realistic scenarios considered in the experimentation. \label{tab:scenarios}}
\begin{tabular}{rrccp{0.40\textwidth}}
\hline
		&				& Number of			& Number of\\
Name 	& Area (m$^2$) 	& entrances (\entrances) 	& objectives (\objectives) & Description\\
\hline 
\textbf{map$_1$} & $160 \times 160$ & 22 	& 61 & A large square surrounded by buildings
										   and with only a few entrances, in which 
										   several monuments and other obstacles are scattered.\vspace{1mm}\\
\textbf{map$_2$} & $224 \times 224$ & 16 	& 25 & An old-town area featuring small plazas 
										   and curvy, non-perpendicular streets that connect them.\vspace{1mm}\\
\textbf{map$_3$} & $352 \times 352$ & 11 	& 9	& A mixture of the two previous scenarios
										   that includes a large square and some adjacent streets\\
\hline
\end{tabular}
\end{table}

\begin{figure}[!t]
\subfloat[][\label{fig:real:HCvsGreedyRadius}]{\adjustbox{trim={.17\width} {.2\height} {.1\width} {.2\height},clip}
{\includegraphics[angle=0,width=0.7\textwidth]{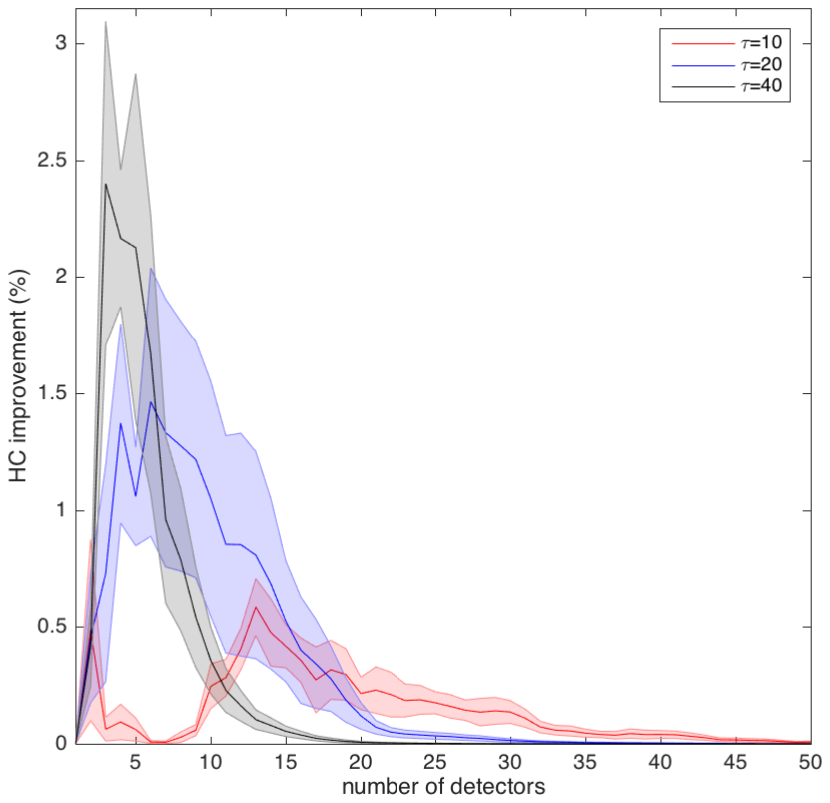}}}~~~~
\subfloat[][\label{fig:real:HCvsGreedyCDF}]{\adjustbox{trim={.17\width} {.2\height} {.1\width} {.2\height},clip}
{\includegraphics[angle=0,width=0.7\textwidth]{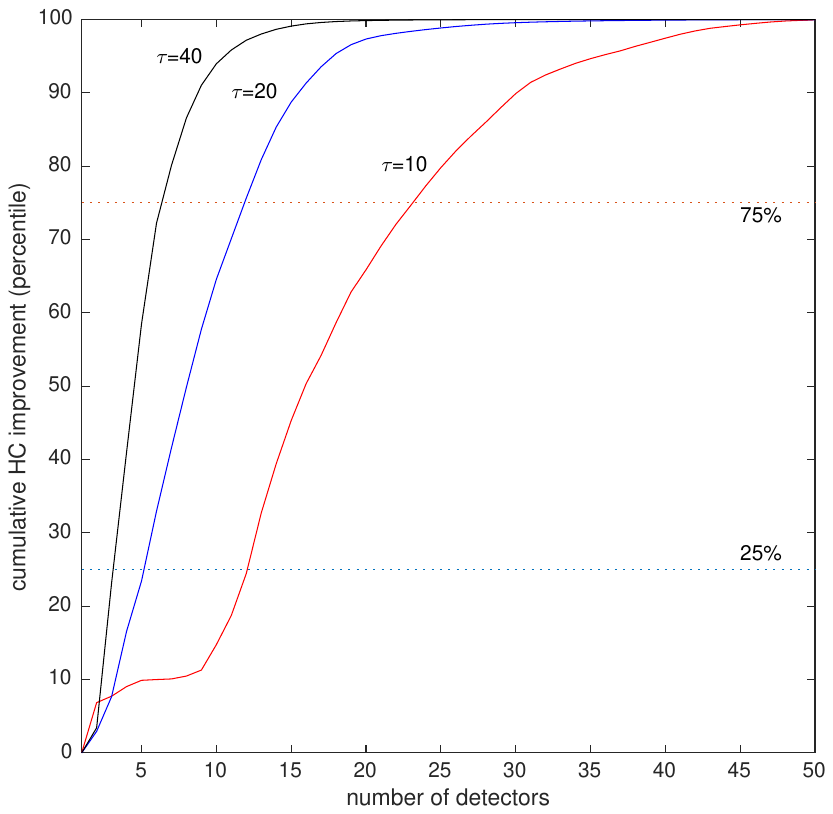}}}
\caption{(a) Relative improvement of solutions provided by HC over those provided by \Greedy{} algorithm 
across the three scenarios, as a function of the number of detectors \detectors\ for different values of the 
detection radius \radius. The shaded areas indicate the standard error. (b) Cumulative distribution of 
improvements depending on the detection radius  \radius. The inner quartiles are shown.}
		\label{fig:real:HCvsGreedy} 
\end{figure}

The experimentation has been done with a number of detectors \detectors{} ranging from 1 to 50
and for different values of the detector radius $\radius\in\{10, 20, 40\}$ meters. 
All experiments have been replicated 25 times for each combination of map, number of detectors
and detection radius. We let the \HC{} algorithm run for 10 seconds or until an local optimum was found (cf. \refAlg{fig:alg:HC}). This resulted in \HC{} taking an average of 4.45 seconds to provide its final solution (standard error $\sigma_{\bar x}=0.043$s).  
\HC{} has consistently found the same solution in each case (\Greedy{}
is deterministic and therefore only one run is required per test case).
The results are summarized in \refFig{fig:real:HCvsGreedyRadius}, in which the average 
relative improvement of \HC{} over \Greedy{} is shown. As can be seen, this improvement 
is always positive for any number of detectors. Indeed, this improvement is statistically 
significant ($p$-value $\approx 0$) according to a Wilcoxon signed rank test for any scenario 
and detection radius, except for \textbf{map}$_3$ and $\radius=40$. Aggregating the results 
for all values of \radius, the improvement is also statistically significant for all scenarios 
(again $p$-value $\approx 0$). Not surprisingly, the relative improvement initially increases 
when more detectors are used, and then declines after reaching a certain peak. Obviously, 
as more and more detectors are placed the fitness function starts to saturate (recall 
\refEq{eq:fitness} -- when the coverage of the paths is large, an additional increase in 
this coverage only implies a small fitness increase due to the exponential decay of the last term 
of the equation). It is nevertheless interesting to note two qualitative differences in behavior 
depending on the value of the detection radius. Firstly, as \radius\ decreases, the peak is 
located in an interval comprising more detectors. This is more precisely depicted in 
\refFig{fig:real:HCvsGreedyCDF}, in which the cumulative distribution of improvement is 
shown (normalized in terms of percentiles for comparison purposes). If we focus on the 
central half of the distribution (that is, the second and third quartiles), it comprises 
$\delta\in[4,7]$ for $\radius=40$m,  
$\delta\in[6,12]$ for $\radius=20$m, and  $\delta\in[13,24]$ for $\radius=10$m. This can 
be interpreted in terms of the area covered by each detector in that case: as the
detection radius increases, larger segments of the paths can be covered and therefore
the fitness functions starts to saturate at an earlier point than it would for smaller
detection radius. In the latter case, adjusting the location of detectors result in
comparatively smaller length variations in the paths covered, implying that the fitness
function saturates more slowly (and hence for a larger interval of detectors) and 
relative improvements are comparatively smaller than for larger values of the 
detection radius. Indeed, if we perform a head-to-head comparison among the values 
obtained on each map and number of detectors for each value of $\radius$, 
we observe that $\radius=10$ ranks better than $\radius\in\{20,40\}$ (mean ranks are 
1.48, 1.87 and 2.65 respectively), the differences being statistically significant 
(Friedman test $p$-value $\approx 0$, Shaffer's test passed at $\alpha=0.05$ for
all pairs of values of \radius). This is in agreement with the interpretation laid down
in \refSect{sect:Sensitivity:Analysis}
for random instances, and underlines the larger interval for which improvements
are obtained for decreasing values of the detector radius. 

\begin{figure}[!t]
	\subfloat[][\label{fig:map2D6R40greedy}]{\adjustbox{trim={0\width} {0\height} {0\width} {0\height},clip} 
		{
			\tikz[baseline=(a.north)]\node[inner sep=0,outer sep=0](a){\includegraphics[angle=0,scale=0.8]{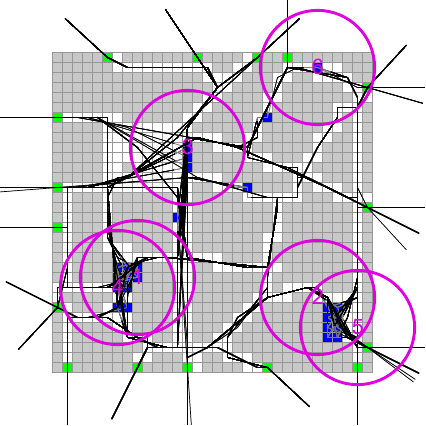}};
		}
	}
	\subfloat[][\label{fig:map2D6R40hc}]{\adjustbox{trim={0\width} {0\height} {0\width} {0\height},clip} 
		{
			\tikz[baseline=(a.north)]\node[inner sep=0,outer sep=0](a){\includegraphics[angle=0,scale=0.8]{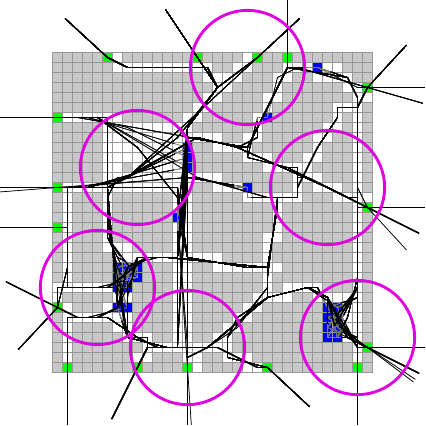}};
		}
	}
	\caption{Solutions found for \textbf{map}$_2$ using $\delta=6$ detectors of radius $\radius=40$m. 
		(a) \Greedy{} (b) \HC{}\label{fig:map2D6R40}. The numbers close to detector locations in 
		the left image indicate the order in which \Greedy{} places the detectors.}
\end{figure}

\refFig{fig:map2D6R40} show a comparison of the solutions found by \Greedy{} and \HC{} 
on a specific case, namely the old-town scenario with $\delta=6$ and $\radius=40$m. 
Notice how the solution created by \Greedy{} places the first three detectors trying to 
cover the densest target areas. Then, it tends to place the next two detectors in 
partial overlap with the first two, aiming to maximize the coverage of these areas by 
virtue of the independent functioning of the detectors. Finally, the last detector is placed
close to a corner of the map, covering paths originating in two entrances. On the other
hand, the solution provided by \HC{} has three detectors in locations rather close,
but not identical, to the first three detectors placed by \Greedy{}. This slight variation
in their locations allows to place the remaining three in strategical positions to cover
all paths originating in the four upper entrances, in three of the lower entrances, and 
in one of the entrances in the right part whose paths were receiving less attention 
by \Greedy. Overall, the solution found by \HC{} has a more global rationale and looks
more spatially balanced.

\section{Conclusions}
\label{sect:conclusions}

Suicide bombing is an infamous form of terrorism that presently causes a large number of casualties worldwide. 
In this work we have researched the problem of placing a number of non-fully reliable detectors on a threat area 
with known targets with the aim of maximizing chances of detecting a suicide-bombing attack, thus minimizing 
the expected number of casualties. 

Apart from a branch-and-bound algorithm that does not scale well with problem size, the only previous 
proposal in the literature to tackle this problem was a greedy algorithm. We have approached this problem 
for the first time to the best of our knowledge from the point of view of iterative heuristics and metaheuristic 
techniques. To this end, we firstly performed an experimental comparison of the different proposed 
algorithms on a benchmark comprising a large set of random instances with different parameterizations. 
Results indicate that, among all considered techniques, a \HC{} algorithm obtains the best results in a
consistent way, clearly outperforming the previous proposal from the literature. Secondly, we did a 
sensitivity analysis in order to determine those settings of problem parameters leading to harder-to-solve 
instances. Thirdly, the \HC{} heuristic (as the best performing algorithm) was experimentally compared 
to the greedy algorithm (as the previous approach in the literature) on a set of real world scenarios 
that could be subject to terrorist attacks. Results of these tests corroborate the good performance 
of \HC{} observed before.

Different lines of research are open as future work to the present study. First of all, alternative algorithmic 
approaches can be explored. For instance, we have observed that the proposed UMDA is able to 
outperform GRASP in case the maximum allowed execution time is increased. Designing an alternative 
probabilistic model or performing new experimental comparisons with increased allowed execution 
times constitute promising extensions to this paper. Another promising proposal is to hybridize in a 
synergistic way the EA and \HC{} algorithms, as best performing algorithms studied in this work. To 
this end, Memetic Algorithms \citep{Neri2012book} constitute a well established and successful 
framework that can be used for this purpose. Lastly, more sophisticated versions of the problem studied could be 
also tackled in future work, like, for instance, considering more precise models of the functioning
of detectors or making dynamic the locations of targets, so 
that their situation and even their number may vary along time.

\begin{acknowledgements}
We would like to thank Mr. Antonio Hern\'andez Bimbela for his help during the initial stage of this project.
\end{acknowledgements}

\end{document}